\documentclass[final]{cvpr}
\usepackage{times}
\usepackage{epsfig}
\usepackage{graphicx}
\usepackage{amsmath}
\usepackage{amssymb}
\usepackage{mathrsfs}
\usepackage{xcolor}
\usepackage{highlight}
\usepackage{enumitem}
\usepackage{subfigure}
\usepackage{enumitem}
\usepackage{algorithm}
\usepackage{multirow}
\usepackage{balance}
\usepackage{algorithmic}
\usepackage{pifont}
\usepackage{balance}
\usepackage{gensymb}
\usepackage{caption}
\usepackage{balance}
\usepackage[normalem]{ulem}

\usepackage{color, colortbl}

\algsetup{linenosize=\small}
\pdfoutput=1

\usepackage[pagebackref=false,breaklinks=true,letterpaper=true,colorlinks,bookmarks=false]{hyperref}

\newcommand{\cmark}{\ding{51}}%
\newcommand{\xmark}{\ding{55}}%
\newcommand{\formattedparagraph}[1]{\noindent \textbf{#1}}

\def\cvprPaperID{4294} 
\def\confYear{CVPR 2021}
\begin{document}

\title{Uncalibrated Neural Inverse Rendering for Photometric Stereo of General Surfaces}

\author{\quad Berk Kaya$^1$\quad Suryansh Kumar$^1$ \quad Carlos Oliveira$^1$ \quad Vittorio Ferrari$^{2}$\quad Luc Van Gool$^{1, 3}$\\
Computer Vision Lab, ETH Z\"urich${^1}$, Google Research$^2$, KU Leuven$^3$
}

\maketitle

\begin{abstract}
This paper presents an uncalibrated deep neural network framework for the photometric stereo problem. For training models to solve the problem, existing neural network-based methods either require exact light directions or ground-truth surface normals of the object or both. However, in practice, it is challenging to procure both of this information precisely,  which restricts the broader adoption of photometric stereo algorithms for vision application. To bypass this difficulty, we propose an uncalibrated neural inverse rendering approach to this problem. Our method first estimates the light directions from the input images and then optimizes an image reconstruction loss to calculate the surface normals, bidirectional reflectance distribution function value, and depth. Additionally, our formulation explicitly models the concave and convex parts of a complex surface to consider the effects of interreflections in the image formation process. Extensive evaluation of the proposed method on the challenging subjects generally shows comparable or better results than the supervised and classical approaches.
\end{abstract}

\vspace{-0.5cm}
\section{Introduction}
Since Woodham's seminal work \cite{woodham1980photometric}, the photometric stereo problem has become a popular choice to estimate an object's surface normals from its light varying images. The formulation proposed in that paper assumes the Lambertian reflectance model of the object, and therefore, it does not apply to general objects with unknown reflectance property. While multiple-view geometry methods exist to achieve a similar goal \cite{schonberger2016structure, furukawa2009accurate, wu2012schematic, zheng2015structure, kumar2019jumping, hartley2003multiple, kumar2017monocular, kumar2019superpixel}, photometric stereo is excellent at recovering fine details on the surface, like indentations, imprints, and even scratches. Of course, the solution proposed in Woodham's paper has some unrealistic assumptions. Still, it is central to the development of several robust algorithms \cite{wu2010robust, ikehata2012robust, queau2017non, alldrin2008photometric, goldman2009shape, higo2010consensus} and also lies at the core of the current state-of-the-art deep photometric stereo methods \cite{ikehata2018cnn, taniai2018neural, chen2018ps, chen2019self, chen2020deep, logothetis2020px, DBLP:conf/bmvc/LogothetisBMC20, santo2020deep}. 

Generally, deep learning-based photometric stereo methods assume a calibrated setting, where all the light source information is given both at the train and test time \cite{ikehata2018cnn, santo2017deep, chen2018ps, taniai2018neural}. Such methods attempt to learn an explicit relation between the reflectance map and the ground-truth surface normals. But, the exact estimation of light directions is a tedious process and requires expert skill for calibration.  Motivated by that, Chen \etal \cite{chen2019self, chen2020deep} recently proposed an uncalibrated photometric stereo method. Though it estimates light directions using image data, the proposed method requires ground-truth surface normals for training the neural network. Certainly, procuring
ground-truth 3D surface geometry is difficult, if not impossible, which makes the acquisition task of correct surface normals strenuous. For 3D data acquisition, active sensors are mostly used, which is expensive and often needs post-processing of the data to remove noise and outliers. Hence, the necessity of ground-truth surface normals limits the usage of such an approach.

\begin{figure*}[t]
\centering
\subfigure[\label{fig:light_effect_on_tablet_illustration} Photometric Stereo Setup ]{\includegraphics[width=0.40\linewidth, height=0.30\textwidth]{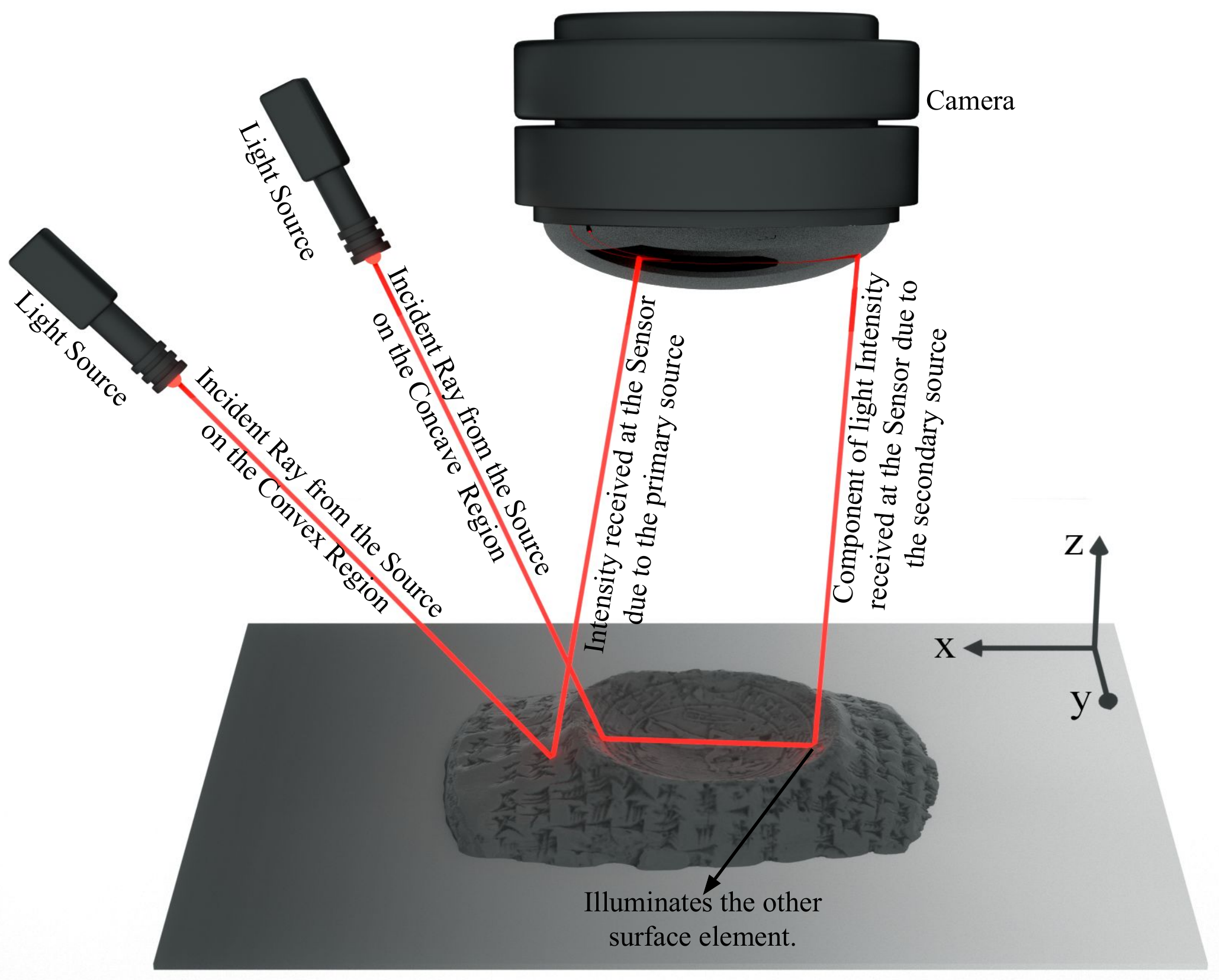}}
\subfigure[\label{fig:vase_comparison} Qualitative and Quantitative Comparison]{\includegraphics[width=0.35\linewidth, height=0.30\textwidth]{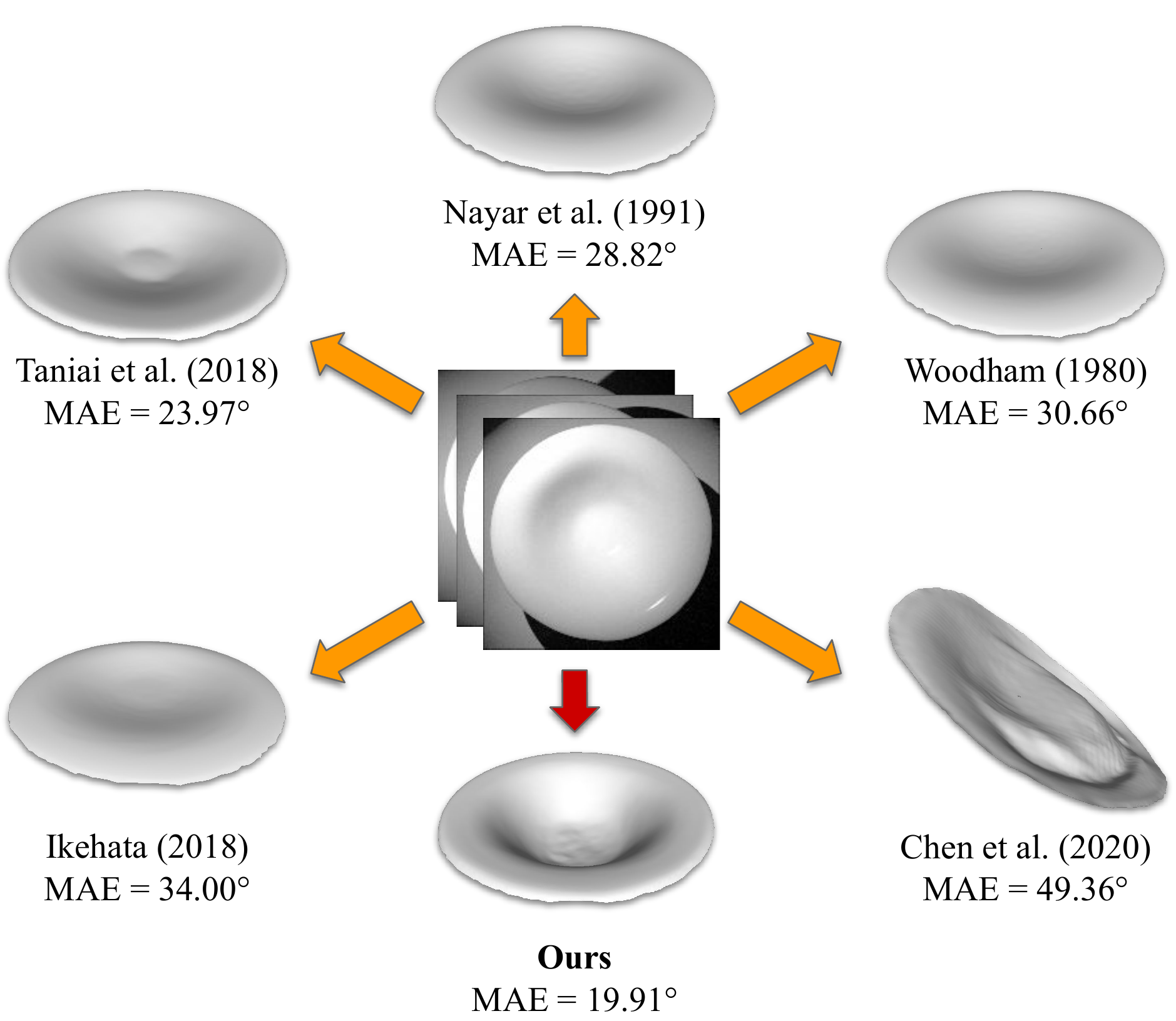}}
\caption{\footnotesize  (a) Example showing the interreflection effect due to concave geometric structure. The light from the primary source hits the concave region of the surface that illuminates the other surface points which then act as a secondary light source. (b) Comparison of our approach against the classical and deep-learning methods on the Vase dataset which shows that it performs better than others. We used Mean Angular Error (MAE) metric to report the results. }
\label{fig:intereflection_phenomena_result_comparison}
\end{figure*}

Further, most photometric stereo methods, including current deep-learning methods, assume that each surface point is illuminated only by the light source, which generally holds for a convex surface \cite{nayar1991shape}. However, objects, mainly from ancient architectures, have complex geometric structures, where the shape may compose of convex, concave, and other fine geometric primitives (see Fig.\ref{fig:light_effect_on_tablet_illustration}). When illuminated under a varying light source, certain concave parts of the surface might reflect light onto other parts of the object, depending on its position. Surprisingly, this phenomenon of interreflections is often ignored in the modeling and formulation of a photometric stereo problem, despite its vital role in the object's imaging \cite{ikehata2018cnn, taniai2018neural, chen2018ps, chen2019self, chen2020deep}.

In this work, we overcome the above shortcomings by proposing an uncalibrated neural inverse rendering network. We first estimate all the light source directions and intensities using image data. Computed light source information is then fed into the proposed neural inverse rendering network to estimate the surface normals. The idea is, those correct surface normals, when provided to the rendering equation, should reconstruct the input image as close as possible. Consequently, we can bypass the requirement of the ground-truth surface normals at train time. Unlike recent methods, we model the effects of both the light source and the interreflections for rendering the image.  Although one can handle interreflection using classical methods \cite{nayar1991shape, chandraker2005reflections}, the reflectance characteristics of different types of material are quite diverse. Hence, we want to leverage neural network's powerful capability to learn complex reflectance behavior from the input image data.

For evaluation, we performed experiments on DiLiGenT dataset \cite{shi2016benchmark}. We noticed that the objects present with this dataset are not apt for studying interreflections. To that end, we proposed a novel dataset to study the behavior and effect of interreflections on the object's imaging \S \ref{sec:Experiment_results}. We observed that ignoring interreflections can dramatically affect the accuracy of the surface normals estimate (see Fig \ref{fig:vase_comparison}).
To sum up, our paper makes the following contributions:
\begin{itemize}[leftmargin=*,topsep=0pt, noitemsep]
    \item This paper presents an uncalibrated deep photometric stereo method that does not require ground-truth surface normals at train time to solve photometric stereo.
    \item  Our work considers the contribution of both the source light and interreflections in the image formation process. Consequently, our approach is more general and applicable to a wide range of objects.
    \item The proposed method leverages neural inverse rendering principles to infer the surface normals, depth, and spatially varying bidirectional reflectance distribution function (BRDF) values from input images. Our method generally provides comparable or better results than the classical \cite{nayar1991shape, alldrin2007resolving, shi2010self, wu2013calibrating, lu2013uncalibrated, papadhimitri2014closed, lu2017symps} and the recent supervised uncalibrated deep learning methods \cite{chen2018ps, chen2020neural, chen2020deep}.
\end{itemize}

\section{Related Work}
For comprehensive review on photometric stereo readers may refer to Herbort \etal \cite{herbort2011introduction}, and Chen \etal \cite{chen2020deep} work.

\smallskip
\formattedparagraph{1. Calibrated Photometric Stereo.} The methods proposed under this setting assume that all the light source information is known for computing surface normals. Several calibrated methods have been proposed to handle non-Lambertian surfaces \cite{mukaigawa2007analysis, wu2009photometric, wu2010robust, miyazaki2010median, oh2013partial, ikehata2014photometric}. These methods assume non-Lambertian effects, such as specularities, are sparse and confined to a local region of the surface. So, they filter them before computing surface normals. For example, Wu \etal \cite{wu2010robust} proposed a rank minimization approach to robustify photometric stereo. Oh \etal \cite{oh2013partial} introduced a partial sum of singular values optimization algorithm for the low-rank normal matrix recovery. Other popular outlier rejection methods were based on RANSAC \cite{mukaigawa2007analysis}, Bayesian regression \cite{ikehata2012robust, ikehata2014photometric}, and expectation-maximization \cite{wu2009photometric}.

With the recent success of deep learning in many computer vision areas, several learning-based approaches have also emerged for the photometric stereo problem. Santo \etal \cite{santo2017deep} introduced a deep photometric stereo network (DPSN) that learns the mapping between the surface normals and the reflectance map. Ikehata \cite{ikehata2018cnn} merged all pixel-wise information to an observation map and trained a network to perform per-pixel estimation of normals. In contrast, Taniai \etal \cite{taniai2018neural} used a self-supervised framework to recover surface normals from input images. Yet,  it uses the classical photometric equation that fails to model interreflections. Moreover, it uses Woodham's method \cite{woodham1980photometric} to initialize the surface normals in their loss function which is not robust, and therefore, their trained network model is susceptible to noise and outliers.

\smallskip
\formattedparagraph{2. Uncalibrated Photometric Stereo.}
These methods assume unknown light source information for solving photometric stereo. However, not knowing the light sources leads to an ambiguity \ie, there exists a set of surfaces under unknown distant light sources that can lead to identical images. Hence, the actual surface can be recovered up to a three-parameter ambiguity popularly known as Generalized Bas-Relief (GBR) ambiguity \cite{belhumeur1999bas, chandraker2005reflections}. Existing methods eliminate this ambiguity by making some additional assumptions in their proposed solution. Alldrin \etal \cite{alldrin2007resolving} assumes bounded values on the GBR variables and resolves the ambiguity by minimizing the entropy of albedo distribution. Shi \etal \cite{shi2010self} 
assumes at least four pixels with different normals but the same albedo. Papadhimitri \etal \cite{papadhimitri2014closed} presents a closed-form solution by detecting local diffuse reflectance maxima (LDR). Other methods assume perspective projection \cite{papadhimitri2013new}, specularities \cite{georghiades2003incorporating, drbohlav2005can}, low-rank \cite{sengupta2018solving},  interreflections \cite{chandraker2005reflections} or symmetry properties of BRDFs \cite{tan2007isotropy, wu2013calibrating, lu2017symps}.

Apart from the traditional methods, Chen \etal \cite{chen2018ps} proposed a learning framework (UPS-FCN). This method bypasses the light estimation process and learns a direct mapping between the image and the surface normal. But, the knowledge of the light source would provide useful evidence about the surface normals, and therefore completely ignoring the light source data seems implausible. The self-calibrating deep photometric stereo networks work \cite{chen2019self} recently introduced an initial lighting estimation stage (LCNet) from images to overcome the problem with UPS-FCN. Recently, Chen \etal \cite{chen2020learned} also proposed a guided calibration network (GCNet) to overcome the limitations of LCNet.  Unlike existing uncalibrated deep-learning methods that rely heavily on ground-truth surface normals for training, our method can solve photometric stereo by using an image reconstruction term as a function of estimated surface normals. The goal is to let the network learn the image formation process and the complex reflectance model of the object via explicit interreflection modeling.

\section{Photometric Stereo}
Photometric stereo aims to recover the surface normals of an object from its multiple images captured under varying light illuminations. It assumes a unique point light source per image taken by a camera from a constant view direction $\mathbf{v}$ which is commonly assumed to be at $(0, 0, 1)^{T}$. Under such configuration, when a surface point $\mathbf{x}$ is illuminated by a distant point light source from direction `$\mathbf{l}_{s} \in \mathbb{R}^{3 \times 1}$', the image intensity $X_s(\mathbf{x})$ measured by the camera due to $s^{th}$ source in the view direction $\mathbf{v}$ is given by
\begin{equation}
    \begin{aligned}\label{eq:generalPS}
       {X}_{s}(\mathbf{x}) = {e}_{s}\cdot\rho\big(\mathbf{n}(\mathbf{x}), \mathbf{l}_{s}, \mathbf{v}\big) \cdot \zeta_a\big(\mathbf{n}(\mathbf{x}), \mathbf{l}_{s}\big) \cdot \zeta_c(\mathbf{x})
    \end{aligned}
\end{equation}
Here, the camera projection model is assumed to be orthographic.
The function $\rho(\mathbf{n}(\mathbf{x}),\mathbf{l}_{s},\mathbf{v})$ gives the BRDF value,  $\zeta_a (\mathbf{n}(\mathbf{x}), \mathbf{l}_{s}) = \max(\mathbf{n}(\mathbf{x})^{T}\mathbf{l}_{s}, 0)$ accounts for the attached shadow,
and  $\zeta_c(\mathbf{x}) \in \{0, 1\}$ assign $0$ or $1$ value to $\mathbf{x}$ depending on whether it lies in the cast shadow region or not. ${e}_{s} \in \mathbb{R}_+$ is a scalar for light intensity value, and $\mathbf{n}(\mathbf{x}) \in \mathbb{R}^{3 \times 1}$ is the surface normal vector at point $\mathbf{x}$. Eq:\eqref{eq:generalPS} is most-widely used photometric stereo formulation which generally works well in practise \cite{ chandraker2005reflections, ikehata2012robust, ikehata2018cnn, taniai2018neural, chen2020learned, chen2020deep}. 

\smallskip
\formattedparagraph{1. Classical Photometric Stereo Model.}
It assumes a convex Lambertian surface model resulting in a constant BRDF value across the whole surface.
Additionally, the surface is considered to be illuminated only due to the light source. Under such assumptions, Eq:\eqref{eq:generalPS} becomes a linearly tractable problem and it is possible to recover the surface normals by solving a simple system of linear equations.
Let all the $n$ light source directions be denoted as $\mathbf{L} = [\mathbf{l}_{1}, \mathbf{l}_{2},.., \mathbf{l}_{n}] \in \mathbb{R}^{3 \times n}$ and $m$ unknown surface point normal be $\mathbf{N} = [\mathbf{n(x_1)}, \mathbf{n(x_2)},.., \mathbf{n(x_m)}] \in \mathbb{R}^{3 \times m}$. Using the notation,  we can write Eq:\eqref{eq:generalPS} due to all the light sources and surface points compactly as

\begin{equation}
    \begin{aligned}\label{eq:stdps}
     \mathbf{X}_\textbf{s} = \rho\mathbf{N}^{T} \mathbf{L}
    \end{aligned}
\end{equation}
where, $\mathbf{X}_\textbf{s} \in \mathbb{R}^{m \times n}$ is the matrix consisting of $n$ images with $m$ object pixels stacked as column vectors, and $\rho$ is the constant albedo. The above system can be solved for the surface normals using the matrix pseudo-inverse approach under calibrated setting if $n \geq 3$ (\ie, at least three light sources are given in non-degenerate configuration).

\smallskip
\formattedparagraph{2. Interreflection Model.}
In contrast to the classical photometric stereo, here, the total radiance at a point $\mathbf{x}$  on the surface is the sum of radiance due to light source $s$ and the radiance due to interreflection from other surface points.

\begin{equation}
    \begin{aligned}\label{eq:nayercont}
        X(\mathbf{x}) = \overbrace{{X}_{s}(\mathbf{x})}^{\textrm{due to light source}} + \overbrace{\frac{\rho(\mathbf{x})}{\pi} \int_{\Omega}K(\mathbf{x}, \mathbf{x}')X(\mathbf{x}')d\mathbf{x}'}^{\textrm{due to interreflections}}
    \end{aligned}
\end{equation}
where, $\Omega$ represents the surface, $\mathbf{x}'$ is another surface point, and $d\mathbf{x}'$ the differential surface element at $\mathbf{x}'$. The value of the interreflection kernel `$K$' at $\mathbf{x}$ due to $\mathbf{x}'$  is defined as:
\begin{equation}
    \begin{aligned}\label{eq:nayerkernelmat}
    K(\mathbf{x}, \mathbf{x}') = \Big(\frac{ (\mathbf{n}(\mathbf{x})^{T}(\mathbf{-r}))\cdot(\mathbf{n}(\mathbf{x'})^{T}\mathbf{r})\cdot V (\mathbf{x}, \mathbf{x'})}{ (\mathbf{r}^{T}\mathbf{r})^2 } \Big)
    \end{aligned}
\end{equation}

The values of $K$, when measured for each surface element form a symmetric and positive semi-definite matrix. In Eq:(\ref{eq:nayerkernelmat}), $V(\mathbf{x}, \mathbf{x'}) $ captures the visibility. When $\mathbf{x}$ occludes $\mathbf{x}'$ or vice-versa then $V$ is 0. Otherwise, $V$ gives the orientation between the two points using the following expression:
\begin{equation}
    \begin{aligned}\label{eq:nayerviewmat}
    V (\mathbf{x}, \mathbf{x'}) = 
        \Big( \frac{\mathbf{n}(\mathbf{x})^{T}(-\mathbf{r}) + |\mathbf{n}(\mathbf{x})^{T}(-\mathbf{r})| }{2|\mathbf{n}(\mathbf{x})^{T}(-\mathbf{r})|}\Big)\\
        \cdot \Big( \frac{\mathbf{n}(\mathbf{x'})^{T}\mathbf{r} + |\mathbf{n}(\mathbf{x'})^{T}\mathbf{r}| }{2|\mathbf{n}(\mathbf{x'})^{T}\mathbf{r}|}\Big)
    \end{aligned}
\end{equation}
where, $\mathbf{n}(\mathbf{x})$ and $\mathbf{n}(\mathbf{x}')$ are the surface normal at $\mathbf{x}$ and $\mathbf{x}'$, and $\mathbf{r} = \mathbf{x} - \mathbf{x}'$ is the vector from   $\mathbf{x}'$ to $\mathbf{x}$. Substituting $V$ and $K$ in Eq:\eqref{eq:nayercont} gives an infinite sum over every infinitesimally small surface element (point) and therefore, it is not computationally easy to find a solution to $X(\mathbf{x})$ in its continuous form. Nevertheless, the solution to Eq:\eqref{eq:nayercont} is guaranteed to converge as $\rho(\mathbf{x}) < 1$ for a real surface. To practically implement the interreflection model, the object surface is discretized into $m$ facets \cite{nayar1991shape}.
Assuming the radiance and albedo values to be constant within each facet, then Eq:(\ref{eq:nayercont}) for the $i^{th}$ facet becomes $ X_i = X_{si} + \frac{\rho_i}{\pi} \sum_{j=1, ~j \neq i}^m X_j K_{ij}$, where $X_i \in \mathbb{R}^{n \times 1}$ and $\rho_i$ are the radiance and albedo of facet $i$.  Considering the contribution of all the light sources for each facet, it can be compactly re-written as:

\begin{equation}
    \begin{aligned}\label{eq:labelnayerupdate}
        & \displaystyle \mathbf{X} = \mathbf{X}_\mathbf{s} + \mathbf{P}\mathbf{K}\mathbf{X},  ~~\Rightarrow \mathbf{X} = (\mathbf{I} - \mathbf{P}\mathbf{K})^{-1}\mathbf{X}_\mathbf{s}
    \end{aligned}
\end{equation}
\noindent
where, $\mathbf{X} = [{X}_1, X_2,.,X_m]^{T}$ is the total radiance for all the facets, and $\mathbf{X}_\mathbf{s} = [{X}_{s1}, {X}_{s2},.,X_{sm}]^{T}$ is the light source contribution to the radiance of $m$ facets. Furthermore, $\mathbf{P}$ is a diagonal matrix composed of albedo values and $\mathbf{K}$ is a $m \times m $  interreflection kernel matrix with $\textrm{diag}(\mathbf{K}) = 0$.  Nayar \etal \cite{nayar1991shape} proposed Eq:\eqref{eq:labelnayerupdate} to recover the surface normals for concave objects. The algorithm proposed to estimate surface normals using Eq:\eqref{eq:labelnayerupdate}  first computes the pseudo surface normals by treating the object as directly illuminated by light sources. These pseudo surface normals are then used to iteratively update for the interreflection kernel and surface normals via depth map estimation step, until convergence. In the later part of the paper, we denote the normals estimated using Eq:(\ref{eq:labelnayerupdate}) as $\mathbf{N}_{ny}$. The Nayar's interreflection model assumes Lambertian surfaces and overlooks surfaces with unknown non-Lambertian properties.

\section{Proposed Method}
Given $\mathbf{X}$ = $[{X}_1, X_2,...,X_n]$ a set of $n$ input images and the object mask $\mathbf{O}$, we propose an uncalibrated photometric stereo method to estimate surface normals. Here, each image ${X}_i$ is reshaped as a column vector and not a facet symbol as used in interreflection modeling. Even though the problem with unknown light directions gives rise to the bas-relief ambiguity \cite{belhumeur1999bas}, we leverage the potential of the deep neural networks to learn those source directions from the input image data using a light estimation network \S \ref{subsec:lcnet}. The estimated light directions are used by the inverse rendering network \S \ref{subsec:inverse_rendering_network} to infer the unknown BRDFs and surface normals using our proposed rendering equation. Our rendering approach explicitly utilizes the role of the light source and interreflections in the image reconstruction process. 

\subsection{Light Estimation Network}\label{subsec:lcnet}
\begin{figure}[t]
\subfigure[\label{fig:sphere_illustration} Source Discretization]{\includegraphics[width=0.40\linewidth, height=0.18\textwidth]{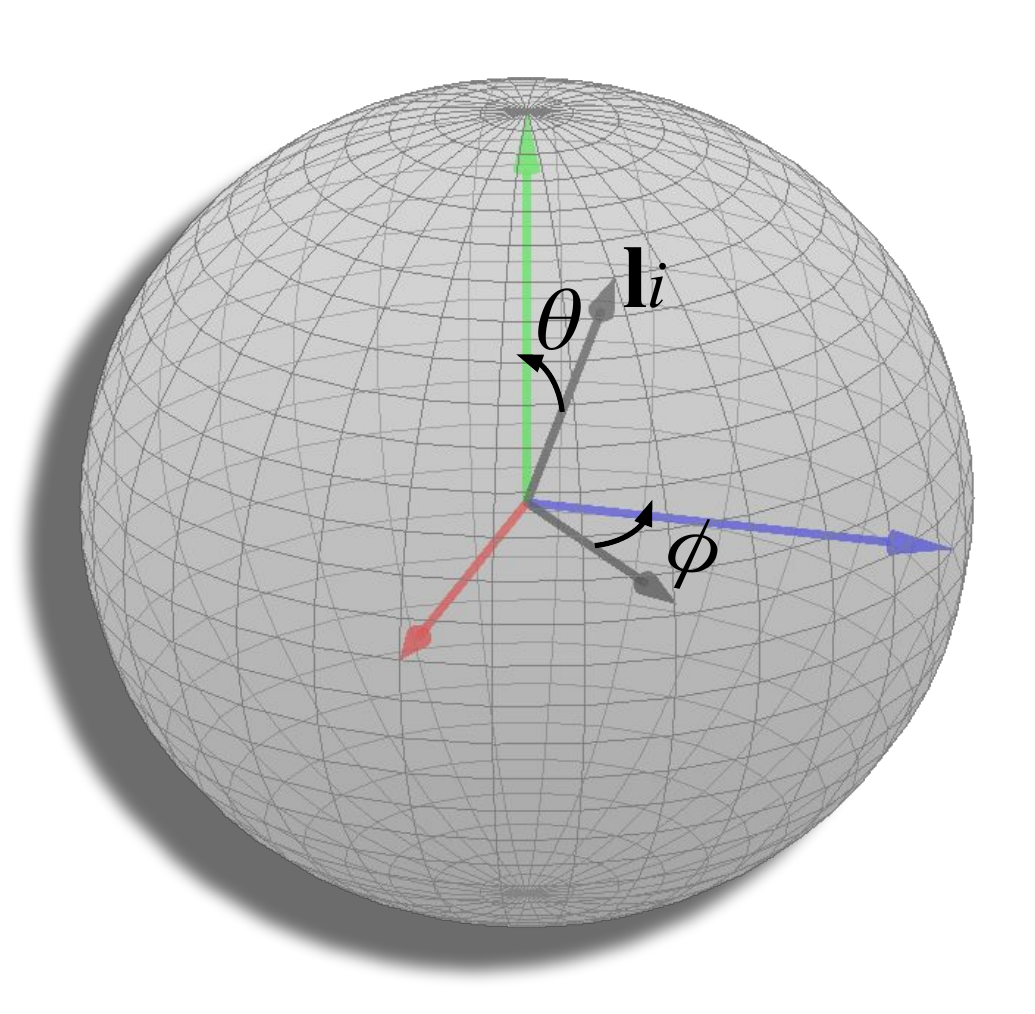}}
~~~~~~~~~~~~\subfigure[\label{fig:radiance_illustration} Surface Reflectance]{\includegraphics[width=0.45\linewidth, height=0.16\textwidth]{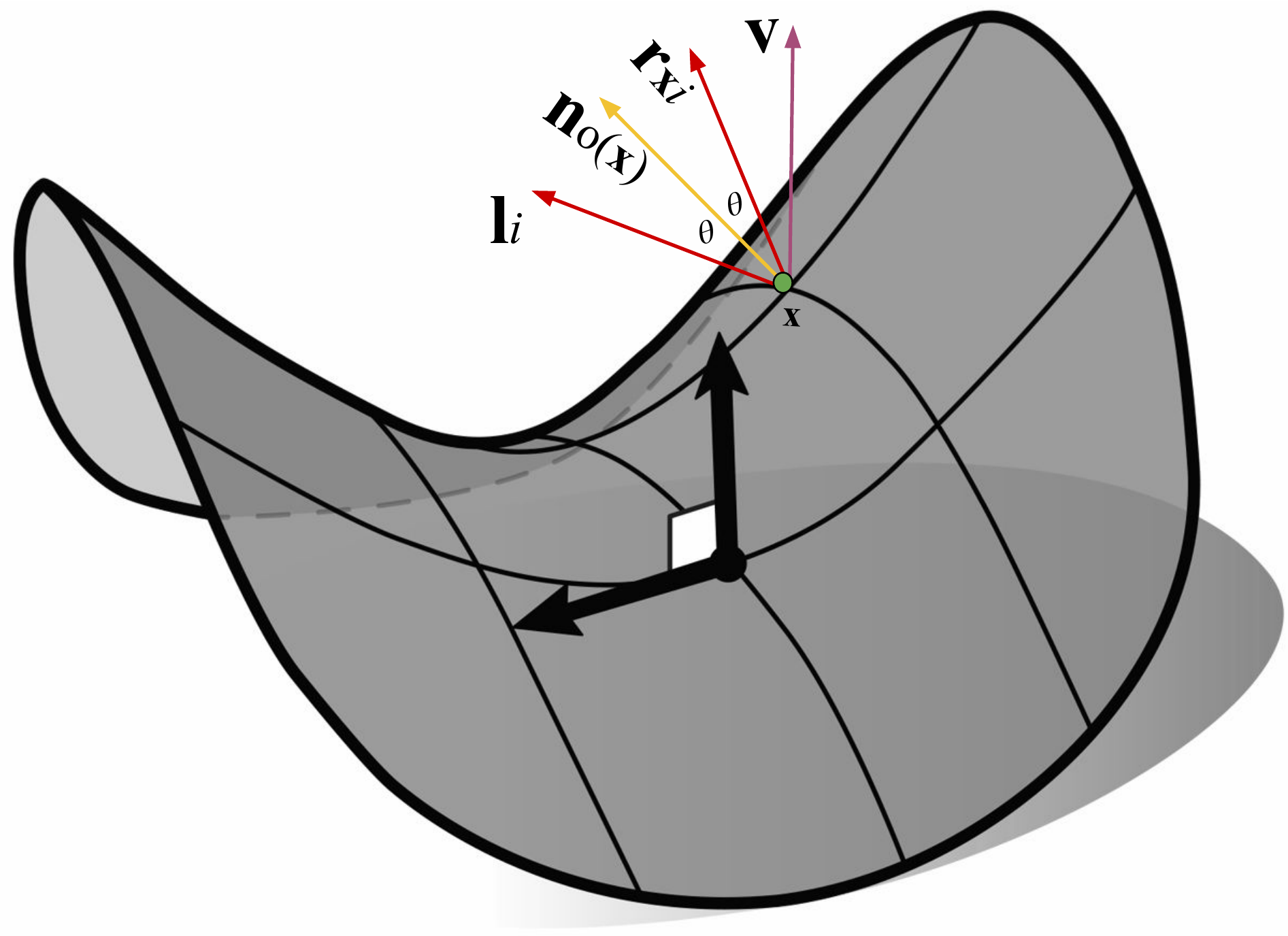}}
\caption{\footnotesize 
(a) The estimated source directions are given by two parameters: $\phi\in[0,\pi]$ and $\theta\in[-\pi/2,\pi/2]$. 
(b) Illustration of surface reflectance. When light ray $\mathbf{l}_{i}$ hits a surface element, the specular component along the view-direction of the point x due to $i^{th}$ source is given by $\mathbf{r}_{\mathbf{x}i}$. Figure 2(b) geometry presentation is inspired by Keenan work \cite{Crane:2013:CGP}.
}
\end{figure}

Given $\mathbf{X}$ and $\mathbf{O}$, the light estimation network predicts the light source intensities ($\mathbf{e}_i$'s) and direction vectors ($\mathbf{l}_i$'s). We can train such a network either by regressing the intensity values and the corresponding unit vector in the source's direction or classifying intensity values into pre-defined angle-range bins. The latter choice seems reasonable as it is easier than regressing the exact direction and intensity values. Further, quantizing the continuous space of directions and intensities for classification makes the network robust to small changes due to outliers or noise. Following that, we express the light source directions in the range $\phi \in [0,\pi]$ for azimuth angles and $\theta \in [-\pi/2,\pi/2]$ for elevation angles (Fig.\ref{fig:sphere_illustration}).  We divide the azimuth and elevation spaces into $K_d = 36$ classes. We classify azimuth and elevation separately, which reduces the problem's dimensionality and leads to efficient computation. Similarly, we divide the light intensity range $[0.2, 2.0]$ into $K_e = 20$ classes \cite{chen2019self}.  

We used seven feature extraction layers to extract image features for each input image separately, where each layer applies $3 \times 3$ convolution and LReLU activation \cite{xu2015empirical}. The weights of the feature extraction layers are shared among all the input images. However, single image features cannot completely disambiguate the object geometry with the light source information. Therefore, we utilize multiple images to have a global implicit knowledge about the surface's geometry and its reflectance property.  We use image specific local features and combine them using a fusion layer to get a global representation of the image set via a max-pooling operation (Fig.\ref{fig:pipeline}). The global feature representation with the image-specific features is then fed to a classifier. The classifier applies four layers of $3 \times 3$ convolution and LReLU activation \cite{xu2015empirical} as well as two fully-connected layers to provide output softmax probability vectors for azimuth ($K_d$), elevation ($K_d$), and intensity ($K_e$). Similar to the feature extraction, the classifier weights are shared among each other. The output value with maximum probability is converted into a light direction vector $\mathbf{l}_i$ and scalar intensity $\mathbf{e}_i$.

\smallskip
\formattedparagraph{Loss function for Light Estimation Network.}
The light estimation network is trained using a multi-class cross-entropy loss \cite{chen2019self}. The total calibration loss $\mathscr{L}_{\textrm{calib}}$ is:
\begin{equation}
    \begin{aligned}\label{eq:lcnet_loss}
        \mathscr{L}_{\textrm{calib}} = \mathscr{L}_{az} +  \mathscr{L}_{el} + \mathscr{L}_{in}
    \end{aligned}
\end{equation}
Here $\mathscr{L}_{az}$, $\mathscr{L}_{el}$, and $\mathscr{L}_{in}$ are the loss terms for azimuth, elevation, and intensity respectively. We used synthetic Blobby and Sculpture datasets \cite{chen2018ps} to train the network.
The light source labels from these datasets are used for supervision at the train time. The network is trained using the above loss for once and the same network is used at the test time for all other datasets \S \ref{sec:Experiment_results}.

\subsection{Inverse Rendering Network}\label{subsec:inverse_rendering_network}
To estimate an object surface normals from $\mathbf{X}$, we leverage neural networks' powerful capability to learn from data. The prime reason for that is, it is difficult to mathematically model the broad classes of BRDFs without any prior assumptions about the reflectance model \cite{georghiades2003incorporating, chung2008efficient, goldman2009shape}. Although there are methods to estimate BRDF values using its isotropic and low-frequency property \cite{ikehata2014photometricisotropic, shi2013bi}, it prohibits the modeling of unrestricted reflectance behavior of the material. Instead of such explicit modeling, we build on the idea of neural inverse rendering \cite{taniai2018neural}, where the BRDFs and surface normals are predicted during the image reconstruction process by the neural network.
We go beyond Taniai \etal \cite{taniai2018neural} work by proposing an inverse rendering network that synthesizes the input images using a rendering equation that explicitly uses interreflections to infer surface normals.


\smallskip
\formattedparagraph{(a) Surface Normal Modeling.} 
We first convert $\mathbf{X}$ into a tensor $\mathcal{X} \in \mathbb{R}^{h \times w  \times nc }$, where $h\times w$ denote the spatial dimensions, $n$ is the number of images, and $c$ is the number of color channels ($c=1$ for grayscale and $c=3$ for color images). $\mathcal{X}$ is then mapped to a global feature map $\Phi$ as follows:
\begin{equation}
    \begin{aligned}\label{eq:xif}
        & \displaystyle \Phi = \xi_f(\mathcal{X}, \mathbf{O}, \Theta_f)
    \end{aligned}
\end{equation}
$\mathbf{O}$ is used to separate the object information from the background. $\xi_f$ is a three layer feed-forward convolutional network with learnable parameter $\Theta_f$. Each layer applies $3 \times 3$ convolution, batch-normalization \cite{ioffe2015batch} and ReLU activation \cite{xu2015empirical} to extract  global feature map $\Phi$.
In the next step, we use $\Phi$ to compute the surface normals. Let $\xi_{n1}$ be the function that converts $\Phi$ into output normal map $\mathbf{N}_o$ via $3 \times 3$ convolution and L2-normalization operation.

\begin{equation}
    \begin{aligned}
        & \displaystyle \mathbf{N}_o = \xi_{n1}(\Phi, \Theta_{n1})
    \end{aligned}
\end{equation}
Here, $\Theta_{n1}$ is the learnable parameter. We used the estimated $\mathbf{N}_o$ to compute $\mathbf{N}_{ny}$ using function $\xi_{n2}$.
\begin{equation}
    \begin{aligned}
        & \displaystyle \mathbf{N}_{ny} = \xi_{n2}(\mathbf{N}_o, \mathbf{P}, \mathbf{K})
    \end{aligned}\label{eq:nayarupdateblock}
\end{equation}
$\xi_{n2}$ requires the interreflection kernel $\mathbf{K}$ and albedo matrix $\mathbf{P}$ as input.  To calculate $\mathbf{K}$, we integrate the $\mathbf{N}_o$ over masked object pixel coordinates $(x, y)$ to obtain the depth map \cite{antensteiner2018review, szeliski2010computer}. Afterward, the depth map is used to infer the kernel matrix $\mathbf{K}$ (see Eq:\eqref{eq:nayerkernelmat}). Once we have $\mathbf{K}$, we employ Eq:\eqref{eq:labelnayerupdate} to compute $\mathbf{N}_{ny}$. Later, $\mathbf{N}_{ny}$ is used in the rendering equation (Eq:\eqref{eq:rendering_equation}) for image reconstruction. 

\smallskip
\formattedparagraph{(b) Reflectance Modeling.} For effective learning of BRDFs, it is important to model the specular component. To incorporate that, we feed a specularity map along with the input image as a channel. Consider the specular-reflection direction $\mathbf{r}_{\mathbf{x}i}$ at a surface element $\mathbf{x}$ with normal $\mathbf{n}_o(\mathbf{x})$ due to the $i^{th}$ light source. We compute $\mathbf{r}_{\mathbf{x}i}$ along the view-direction vector $\mathbf{v}$ using the following relation:

\begin{equation}
    \begin{aligned}
    & \displaystyle \mathbf{r}_{\mathbf{x}i} = \mathbf{v}^{T}\Big(2\big(\mathbf{n}_{o}(\mathbf{x})^{T}\mathbf{l}_i)\cdot\mathbf{n}_{o}(\mathbf{x}\big)-\mathbf{l}_i\Big)
    \end{aligned}
\end{equation}
Here, $\|\mathbf{l}_i\|_2, \|\mathbf{n}_o(\mathbf{x})\|_2,  \|\mathbf{r}_{\mathbf{x}i}\|_2$ are 1 (see Fig.\ref{fig:radiance_illustration}).
Computing $\mathbf{r}_{\mathbf{x}i}$ for all surface points provides the specular-reflection map ${R}_i \in \mathbb{R}^{h \times w \times 1}$. Concatenating $X_i \in \mathbb{R}^{h \times w \times c}$ with ${R}_i$ across channel guides the network to learn complex BRDFs. Thus, we compute feature map ${S}_i$ as:

\begin{equation}
    \begin{aligned}
        & \displaystyle {S}_i = f_{sp}({X}_{i}\oplus{R}_{i}, \Theta_{sp})
    \end{aligned}
\end{equation}

\begin{figure*}[t]
    \centering
    \includegraphics[{width=0.95\linewidth}]{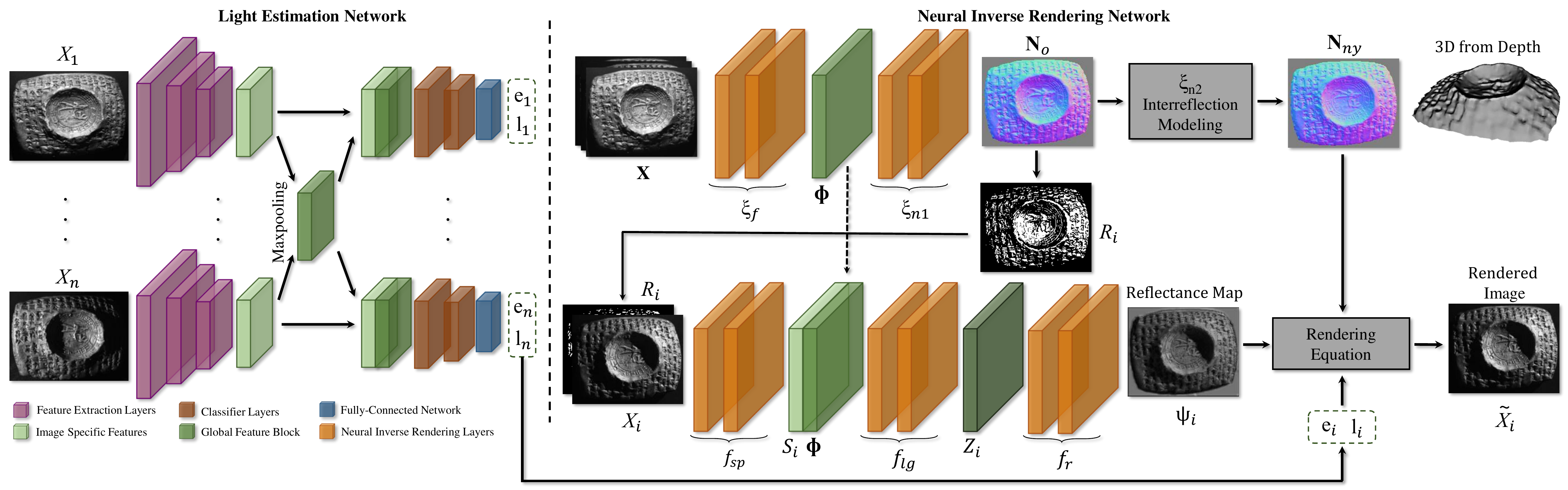}
    \caption{\footnotesize 
The proposed method consists of two networks. Light estimation network initially predicts the light source directions and intensities from input images. Then, neural inverse rendering network uses the images and the light source estimations to recover surface normals, depth and BRDF values. }
    \label{fig:pipeline}
\end{figure*}
We used $\oplus$ to denote the concatenation operation. $f_{sp}$ is a three-layer network where each layer applies $3\times3$ convolution, batch-normalization \cite{ioffe2015batch} and ReLU operations \cite{xu2015empirical}. Although the feature map $S_i$ models the actual specular component of a BRDF, it is computed using a single image observation ${X}_i$ which has limited information. To enrich the feature, we concatenate it with the global features $\Phi$ (see Eq:\eqref{eq:xif}) and compute enhanced feature block ${Z}_i$.

\begin{equation}
    \begin{aligned}\label{eq:zi}
        & \displaystyle {Z}_i = f_{lg}({S}_i\oplus\Phi, \Theta_{lg})
    \end{aligned}
\end{equation}
$f_{lg}$ function applies $1\times1$ convolution, batch normalization \cite{ioffe2015batch} and ReLU operations \cite{xu2015empirical} to estimate ${Z}_i$ . Finally, we define the reflectance function $f_r$ that blends the image specific features with $\Phi$ along with the specular component of the image to compute the reflectance map $\Psi_i$.
\begin{equation}
    \begin{aligned}\label{eq:fr}
        & \displaystyle \Psi_i = f_{r}({Z}_i, \Theta_{ri})
    \end{aligned}
\end{equation}
The function $f_{r}$ applies $3\times3$ convolution, batch normalization \cite{ioffe2015batch}, ReLU operation \cite{xu2015empirical} with an additional $3\times3$ convolution layer to compute $\Psi_i$. The predicted $\Psi_i$ by the network contains the BRDFs and cast shadow information. The specular ($\Theta_{sp}$), local-global ($\Theta_{lg}$), and reflectance image ($\Theta_{ri}$) parameters are learned over SGD iteration by the network. Details about the implementation of above functions, learning and testing strategy are described in \S \ref{sec:Experiment_results}.

\smallskip
\formattedparagraph{(c) Rendering equation.}
Assuming photometric stereo setup, once we have the surface normals, reflectance map, and light source information, we render the input image using the following equation:
\begin{equation}
    \begin{aligned}\label{eq:rendering_equation}
        & \displaystyle \Tilde{{X}}_{i} = \Psi_i  \odot \big({e}_i \cdot \zeta_a (\mathbf{N}_{ny}, \mathbf{l}_i)\big) 
    \end{aligned}
\end{equation}
Here, we explicitly model the effects of interreflections in the image formation. For a given source, $\Psi_i$ encapsulates the BRDF values with the cast shadow information. Further, $\zeta_a$ is defined for the attached shadow. With a slight abuse of notation used in Eq:\eqref{eq:generalPS}, $\zeta_a$ computes the inner product between a light source and the surface normal matrix for each pixel, and the maximum operation is done element-wise \ie, $\max(\mathbf{N}_{ny}^{T}\mathbf{l}_i, 0)$. ${e}_i \in \mathbb{R}_+$ is a scalar intensity value of the light source, and $\odot$ denotes the Hadamard product. Fig.\ref{fig:pipeline} shows the entire rendering network pipeline.

\smallskip
\formattedparagraph{Loss Function for Inverse Rendering Network.}
To train the proposed inverse rendering network, we use $l_1$ loss between the rendered images $\Tilde{\mathbf{X}}$ and input images $\mathbf{X}$ on the masked pixels ($\mathbf{O}$). The network parameters are learned by minimizing the following loss using the SGD algorithm:
\vspace{-0.0cm}
\begin{equation}
    \begin{aligned}\label{eq:reconstruction_loss}
            \mathscr{L}_{rec}(\mathbf{X},  \Tilde{\mathbf{X}}) = \frac{1}{mnc} \sum_{i, c, \mathbf{x}}
            |{X}_{i, c}(\mathbf{x}) - \Tilde{{X}}_{i, c} (\mathbf{x}) |
    \end{aligned}
\end{equation}
Here, $m$ is the number of pixels within $\mathbf{O}$ and $n, c$ are the number of input images and color channels, respectively. The optimization of the above image reconstruction loss function seems reasonable; but, it may provide unstable behavior leading to inferior results. Therefore, we apply weak supervision to the network at the early stages of the optimization by adding a surface normal regularizer in the loss function using an initial normal estimate $\mathbf{N}_{init}$.
Such a strategy guides the network for stable convergence behavior and a better solution to the surface normals. The total loss function is defined as:
\vspace{-0.0cm}
\begin{equation}
    \begin{aligned}
            \mathscr{L}
            = 
            \mathscr{L}_{rec}(\mathbf{X},  \Tilde{\mathbf{X}}) +
            \lambda_w \mathscr{L}_{weak}(\mathbf{N}_{ny},  {\mathbf{N}_{init}})
    \end{aligned}\label{eq:normal_init}
\end{equation}
where, function $\mathscr{L}_{weak}$ is defined as $\mathscr{L}_{weak}(\mathbf{N}_{ny},  {\mathbf{N}_{init}}) = \frac{1}{m} \sum_{\mathbf{x}} \left\| \mathbf{n}_{ny}(\mathbf{x}) - {\mathbf{n}_{init}}(\mathbf{x}) \right\|_2^2$.
Least-square solution of $\mathbf{N}$ in Eq:\eqref{eq:stdps} can provide weak supervision to the network in the early stage of the optimization. However, such initialization may provide undesirable behavior at times. Therefore, we adhere to the robust optimization algorithm on photometric stereo (\S \ref{sssec:robustinit}) to initialize the surface normal in Eq:\eqref{eq:normal_init}. 

\begin{table*}[t]
\scriptsize
\centering
\resizebox{\textwidth}{!}
{
\begin{tabular}{ c | c | r *{10}{|c} | c  }
\hline
\rowcolor[gray]{0.70}
\textbf{Type} &  \textbf{G.T. Normal} &\textbf{Methods}$\downarrow$ $|$ \textbf{Dataset}  $\rightarrow$ & \textbf{Ball} & \textbf{Cat} & \textbf{Pot1} & \textbf{Bear} & \textbf{Pot2}& \textbf{Buddha} &\textbf{Goblet}&\textbf{Reading} & \textbf{Cow} & \textbf{Harvest} & \textbf{Average} \\
\hline
\rowcolor[gray]{0.92}
\hline
\rowcolor[gray]{0.92}
Classical  & \xmark &Alldrin et al.(2007)~\cite{alldrin2007resolving} & 7.27& 31.45& 18.37& 16.81 &49.16 & 32.81 & 46.54 & 53.65 & 54.72 & 61.70 & 37.25 \\
\hline
\rowcolor[gray]{0.92}
Classical  & \xmark &Shi et al.(2010)~\cite{shi2010self} & 8.90 & 19.84 & 16.68 & 11.98 & 50.68 & 15.54 & 48.79 & 26.93 & 22.73 & 73.86 & 29.59 \\
\hline
\rowcolor[gray]{0.92}
Classical  & \xmark &Wu et al.(2013)~\cite{wu2013calibrating} & 4.39 & 36.55 & 9.39 & \cellcolor{red!40}6.42 & 14.52 & 13.19 & 20.57 & 58.96 & 19.75 & 55.51 & 23.93 \\
\hline
\rowcolor[gray]{0.92}
Classical  & \xmark &Lu et al.(2013)~\cite{lu2013uncalibrated} & 22.43 & 25.01 & 32.82 & 15.44 & 20.57 & 25.76 & 29.16 & 48.16 & 22.53 & 34.45 & 27.63 \\
\hline
\rowcolor[gray]{0.92}
Classical  & \xmark &Pap. et al.(2014)~\cite{papadhimitri2014closed} & 4.77 & 9.54 & 9.51 & 9.07 & 15.90 & 14.92 & 29.93 & 24.18 & 19.53 & 29.21 & 16.66 \\
\hline
\rowcolor[gray]{0.92}
Classical  & \xmark &Lu et al.(2017)~\cite{lu2017symps} & 9.30 & 12.60 & 12.40 & 10.90 & 15.70 & 19.00 & 18.30 & 22.30 & 15.00 & 28.00 & 16.30 \\
\hline
NN-based  & \cmark &Chen et al.(2018)~\cite{chen2018ps} & 6.62 & 14.68 & 13.98 & 11.23 & 14.19 & 15.87 & 20.72 & 23.26 & 11.91 & 27.79 & 16.02 \\
\hline
NN-based  & \cmark &Chen et al.(2018)$^{\dagger}$ ~\cite{chen2018ps}& 3.96 & 12.16 & 11.13 & 7.19 & 11.11 & \cellcolor{red!40}{13.06} & 18.07 & 20.46 & 11.84 & 27.22 & 13.62 \\
\hline
NN-based  & \cmark & Chen et al.(2019)~\cite{chen2019self} & \cellcolor{red!20}{\textbf{2.77}} & \cellcolor{red!40}{8.06} & \cellcolor{red!20}{\textbf{8.14}} & 
{6.89} & \cellcolor{red!20}{\textbf{7.50}} & \cellcolor{red!20}{\textbf{8.97}} & \cellcolor{red!20}{\textbf{11.91}} & \cellcolor{red!20}{\textbf{14.90}} & \cellcolor{red!20}{\textbf{8.48}} & \cellcolor{red!20}{\textbf{17.43}} & \cellcolor{red!20}{\textbf{9.51}}\\
\hline
NN-based & \xmark & \textbf{Ours} & \cellcolor{red!40}{3.78} & \cellcolor{red!20}{\textbf{7.91}} & \cellcolor{red!40}{8.75}  & \cellcolor{red!20}{\textbf{5.96}} & \cellcolor{red!40}{10.17} & 13.14 & \cellcolor{red!40}{11.94} & \cellcolor{red!40}{18.22} & \cellcolor{red!40}{10.85} & \cellcolor{red!40}{25.49} & \cellcolor{red!40}{11.62} \\
\hline
\end{tabular}} 
\caption{ \footnotesize Without using ground-truth light or surface normals of this dataset at train time, our method supplies results that is comparable to the recent state-of-the-art \cite{chen2019self}. The $1^{st}$ and $2^{nd}$ best performing methods are colored in light-red and dark-red respectively. \textbf{G.T. Normal} column indicates the use of ground-truth normal at train time. Comparisons are done against well-known uncalibrated methods. $^{\dagger}$ indicates the deeper version of the UPS-FCN model.}
\label{tab:diligentuncalibrated}
\end{table*}

\section{Dataset Acquisition and Experiments}\label{sec:Experiment_results}
We performed evaluations of our method on DiLiGenT dataset \cite{shi2016benchmark}. DiLiGenT is a standard benchmark for photometric stereo, consisting of ten different real-world objects. Despite it provides surfaces of diverse reflectances, the subjects are not elegant for studying interreflections. Therefore, we propose a new dataset that is apt for analyzing such complex imaging phenomena. The acquisition is performed using two different setups. In the first setup, we designed a physical dome system to capture the cultural artifacts. It is a $35cm$ hemispherical structure with 260 LEDs on the nodes for directed light projection, and with a camera on top, looking down vertically. The object under investigation lies at the center. Using it, we collected images of three historical artifacts (\textit{Tablet1}, \textit{Tablet2}, \textit{Broken Pot}) with spatial resolution of $180 \times 225$. Ground-truth normals are acquired using active sensors with post-refinements. We noted that it is onerous to capture 3D surfaces with high-precision. For this reason, we simulated the dome environment using Cinema 4D software with 100 light sources. Using this synthetic setup, we rendered images of three objects (\textit{Vase}, \textit{Golf-ball}, \textit{Face}) with spatial resolution of $256 \times 256$.  
Our dataset introduces new subjects with general reflectance property to initiate a broader adaptation of photometric stereo algorithm for extracting 3D surface information of real objects.

\smallskip
\formattedparagraph{Implementation Details.} 
Our method is implemented in PyTorch \cite{paszke2017automatic}. The light estimation network is trained using Blobby and Sculpture datasets \cite{chen2018ps} with Adam \cite{DBLP:journals/corr/KingmaB14} optimizer and initial learning rate of $5 \times 10^{-4}$. We trained the model for 20 epochs with a batch size of 32. The learning rate is divided by two after every 5 epochs. Training of the neural inverse rendering network is not required as it learns the network parameters at the test time. However, the initialization of the network is crucial for stable learning.

\begin{figure}
    \centering
    \includegraphics[{width=0.90\linewidth}]{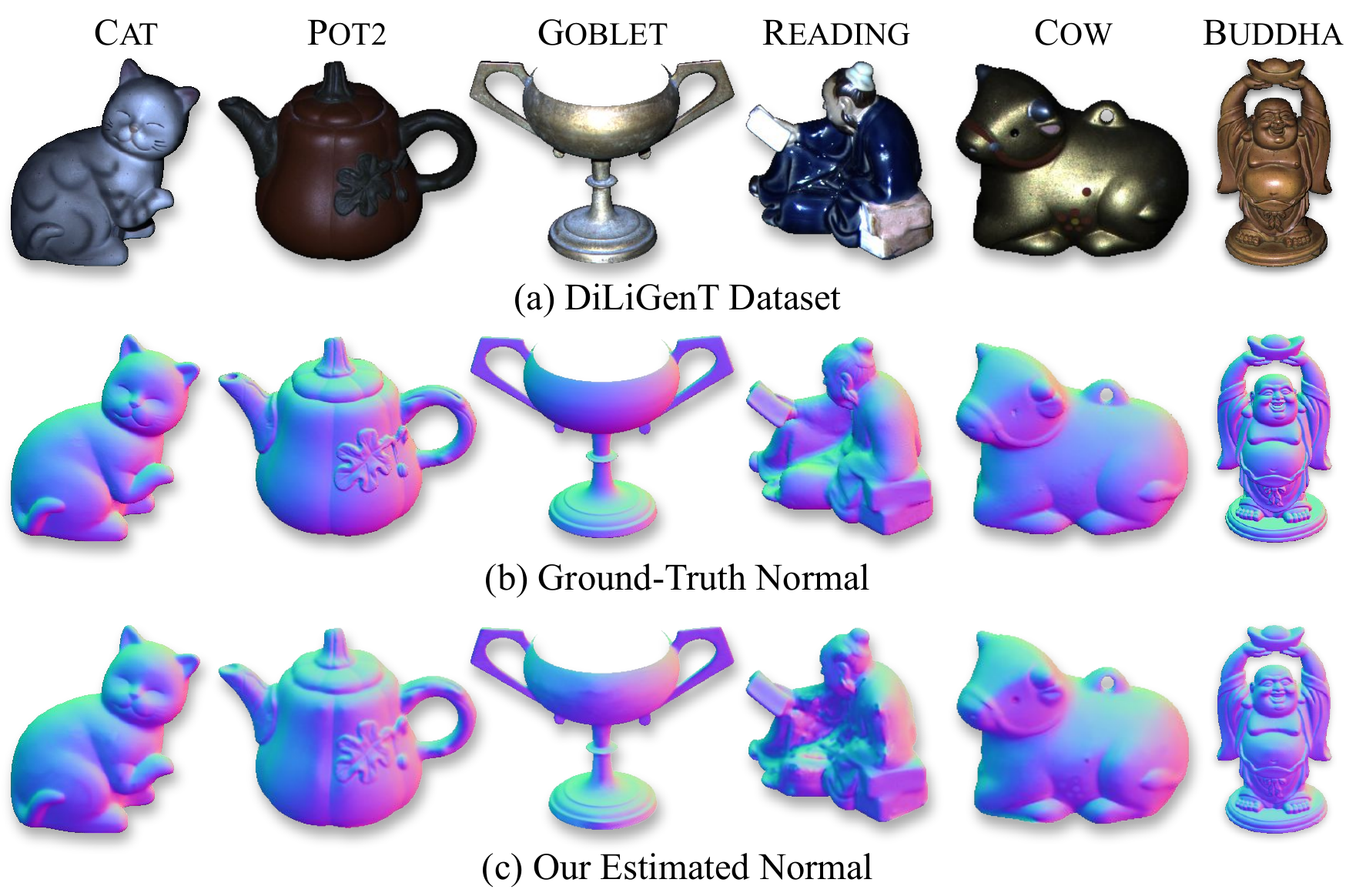}
    \caption{\footnotesize \textbf{Qualitative results on DiLiGenT dataset using our method.} }
    \label{fig:diligent_results}
\end{figure}

\noindent
\textit{$\bullet$  \uline{Initialization}:}
\label{sssec:robustinit}
Our method uses an initial surface normals prior $\mathbf{N}_{init}$ (Eq:\eqref{eq:normal_init})  to warm up the rendering network and to initialize the interreflection kernel $\mathbf{K}$ values. Woodham's classical method \cite{woodham1980photometric} is a conventional way to do so under given light sources. However, initialization using Woodham's method is observed to provide a unstable network behavior leading to inferior results \cite{taniai2018neural}. Therefore, for initialization, we propose to use partial sum of singular values optimization \cite{oh2013partial}. Let $\mathbf{X} \in \mathbb{R}^{m \times n}$, $\mathbf{L} \in \mathbb{R}^{3 \times n}$, $\mathbf{N} \in \mathbb{R}^{3 \times m}$, then Eq:\eqref{eq:stdps} under Lambertian assumption with $\rho=1$ can be written as $ \mathbf{X} = \mathbf{N}^{\mathrm{T}}\mathbf{L} + \mathbf{E}$. Here, $\mathbf{E} \in \mathbb{R}^{m \times n}$ is a matrix of outliers and assumed to be sparse \cite{wu2010robust}. Substituting $\mathbf{Z} = \mathbf{N}^{\mathrm{T}}\mathbf{L}$, the normal estimation under low rank assumption can be formulated as a RPCA problem \cite{wu2010robust}. We know that RPCA performs the nuclear norm minimization of $\mathbf{Z}$ matrix which not only minimizes the rank but also the variance of $\mathbf{Z}$ within the target rank. Now, for the photometric stereo model, it is easy to infer that $\mathbf{N}$ lies in a rank-3 space.  As the true rank for $\mathbf{Z}$ is known from its construction, we do not minimize the subspace variance within the target rank ($K$).  We preserve the variance of information within the target rank while minimizing other singular values outside it via the following optimization:
\begin{equation}\label{eq:pami15ps1}
\begin{aligned}
\centering
& \displaystyle \underset{\mathbf{Z}, \mathbf{E}} {\textrm{min.}} ~\|\mathbf{Z}\|_{r=K} + \lambda\|\mathbf{E}\|_1, ~~\textrm{subject to:}~\mathbf{X} = \mathbf{Z} + \mathbf{E} 
\end{aligned}
\end{equation}
Eq:\eqref{eq:pami15ps1} is a well-studied problem and we solved it using ADMM \cite{boyd2011distributed, oh2013partial, lin2010augmented}. We use the Augmented Lagrangian form of Eq:\eqref{eq:pami15ps1} to solve $\mathbf{Z}$, $\mathbf{E}$ for $K=3$. The recovered solution is used to initialize the surface normal in Eq:\eqref{eq:normal_init}. For detailed derivations, refer to supplementary material.

\begin{figure}
    \centering
    \includegraphics[{width=0.92\linewidth}]{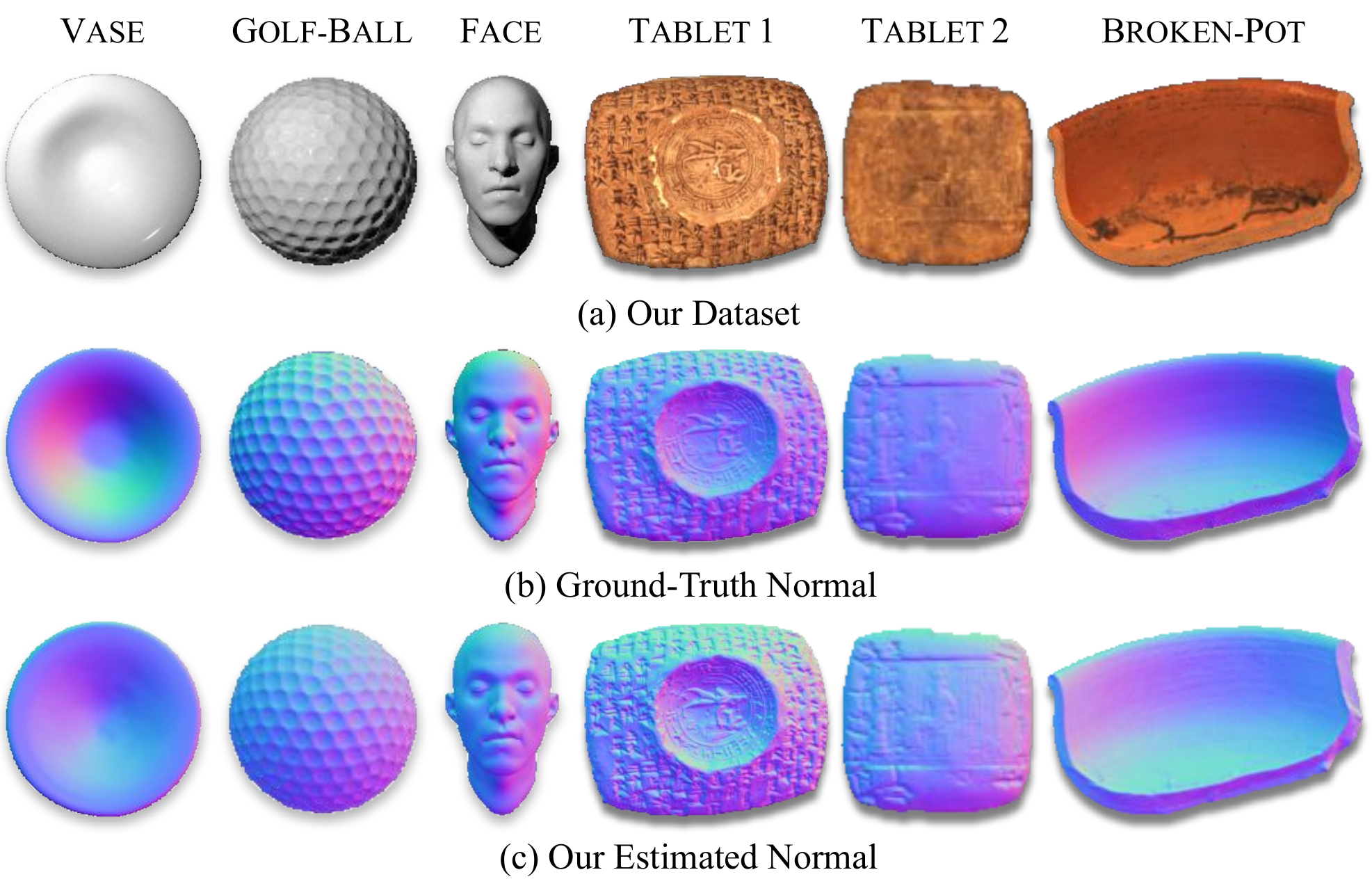}
    \caption{\footnotesize \textbf{Qualitative results of our method on proposed dataset.} }
    \label{fig:our_results}
\end{figure}

\begin{table*}[t]
\scriptsize
\centering
{
{\begin{tabular}{ c | c | r *{6}{|c} | c  }
\hline
\rowcolor[gray]{0.85}
\textbf{Type} &  \textbf{G.T. Normal} & \textbf{Methods}$\downarrow$ $|$ \textbf{Dataset} $\rightarrow$  & \textbf{Vase} & \textbf{Golf-ball}  &  \textbf{Face}   &  \textbf{Tablet 1} & \textbf{Tablet 2} & \textbf{Broken Pot} &   \textbf{~~Average~~} \\
\hline
Classical & \xmark & Nayar et al.(1991)~\cite{nayar1991shape} &  \cellcolor{red!40}28.82 &  \cellcolor{red!40}11.30 &  13.97 &   \cellcolor{red!40}19.14 & 16.34 &  19.43 & \cellcolor{red!40}18.17  \\
\hline
NN-based & \cmark & Chen et al.(2018)~\cite{chen2018ps} & 35.79  &  36.14     &  48.47 &  19.16 &  \cellcolor{red!20}{\textbf{10.69}} & 24.45 & 29.12 \\
\hline
NN-based & \cmark & Chen et al.(2019)~\cite{chen2019self} &  49.36 & 31.61 &  \cellcolor{red!40}13.81 &  16.00 & 15.11 &  \cellcolor{red!20}{\textbf{18.34}} &  24.04 \\
\hline
NN-based & {\xmark} & \textbf{Ours} &  \cellcolor{red!20}\textbf{19.91}  &  \cellcolor{red!20}\textbf{11.04} &   \cellcolor{red!20}\textbf{13.43} &
\cellcolor{red!20}\textbf{12.37} &  \cellcolor{red!40}13.12 &  \cellcolor{red!40}{18.55}  & \cellcolor{red!20}\textbf{14.74}  \\
\hline
\end{tabular}}}
\caption{\footnotesize Comparison against recent uncalibrated deep photometric stereo methods and Nayar \etal \cite{nayar1991shape} on our dataset. In contrast to our approach, Chen \etal \cite{chen2018ps} and Chen \etal \cite{chen2019self} require ground-truth normal for training the network. We can observe that our method shows consistent behavior over a diverse dataset that is on average better than other methods. The two best-performing methods are shaded with light-red and dark-red color respectively.}
\label{tab:uncalibratedmethods}
\end{table*}

\noindent
\textit{$\bullet$ \uline{Testing}:} For testing, we first feed the test images to the light estimation network to get source directions and intensities. For objects like \textit{Vase}, where the cast shadows and interreflections play a vital role in the object's imaging, light estimation network can have questionable behavior. So, we use the light source directions and intensities estimated from a calibration sphere for testing our synthetic objects. Once normal is initialized using our robust approach, we learn inverse rendering network's parameters by minimizing $\mathscr{L}$ of Eq:\eqref{eq:normal_init}. To compute $\mathscr{L}_{rec}$, we randomly sample $10 \% $ of the pixels in each iteration and compute it over these pixels to avoid local minimum. To provide weak-supervision, we set $\lambda _w = \mathcal{L}_{rec} (0 , \mathbf{X})$ to balance the influence of $\mathscr{L}_{rec}$ and $\mathscr{L}_{weak}$ to network learning process. Note that $\lambda _w$ is set to zero after 50 iterations to drop early stage weak-supervision. We perform 1000 iterations in total with initial learning rate of $8 \times 10^{-4}$. The learning rate is reduced by factor of 10 after 900 iterations for fine-tuning. 
Before feeding the images to the normal estimation network, we normalize them using a global scaling constant $\sigma$, \ie the quadratic mean of pixel intensities $\mathbf{X}' = \mathbf{X} / (2 \sigma)$. During the learning of inverse rendering network, we repeatedly update the kernel $\mathbf{K}$ using $\mathbf{N}_{o}$ after every 100 iterations.

\subsection{Evaluation, Ablation Study and Limitation}
\formattedparagraph{(a) DiLiGent Dataset.}
Table(\ref{tab:diligentuncalibrated}) provides statistical comparison of our method against other uncalibrated methods on DiLiGenT benchmark. We used popular mean angular error (MAE) metric to report the results. It can be inferred that our method achieves competitive results on this benchmark with an average MAE of $11.62$ degrees, achieving the second best performance overall without ground-truth surface normal supervision. On the contrary, the best performing method \cite{chen2019self} uses ground-truth normals during training, and therefore, it performs better for objects like \textit{Harvest}, where imaging is deeply affected by discontinuities.

\smallskip
\formattedparagraph{(b) Our Dataset.}
Table(\ref{tab:uncalibratedmethods}) compares our method with other deep uncalibrated methods on the proposed dataset. For completeness, we analyzed Nayar \etal~\cite{nayar1991shape} algorithm by using light sources data obtained using our approach. The results show that our method achieves the best performance overall. We observed that other deep learning methods cannot handle objects like \textit{Vase} as they fail to model complex reflectance behavior. Similarly, Nayar \etal~\cite{nayar1991shape} results indicate that modeling interreflections alone is not sufficient. Since we not only model the effects of interreflections, but also the reflectance mapping associated with the geometry, our method consistently performs well.

\smallskip
\formattedparagraph{(c) Ablation Study.} 
For this study, we validate the importance of robust initialization and interreflection modeling.

\begin{figure}
    \centering
    \includegraphics[{width=1.0\linewidth}]{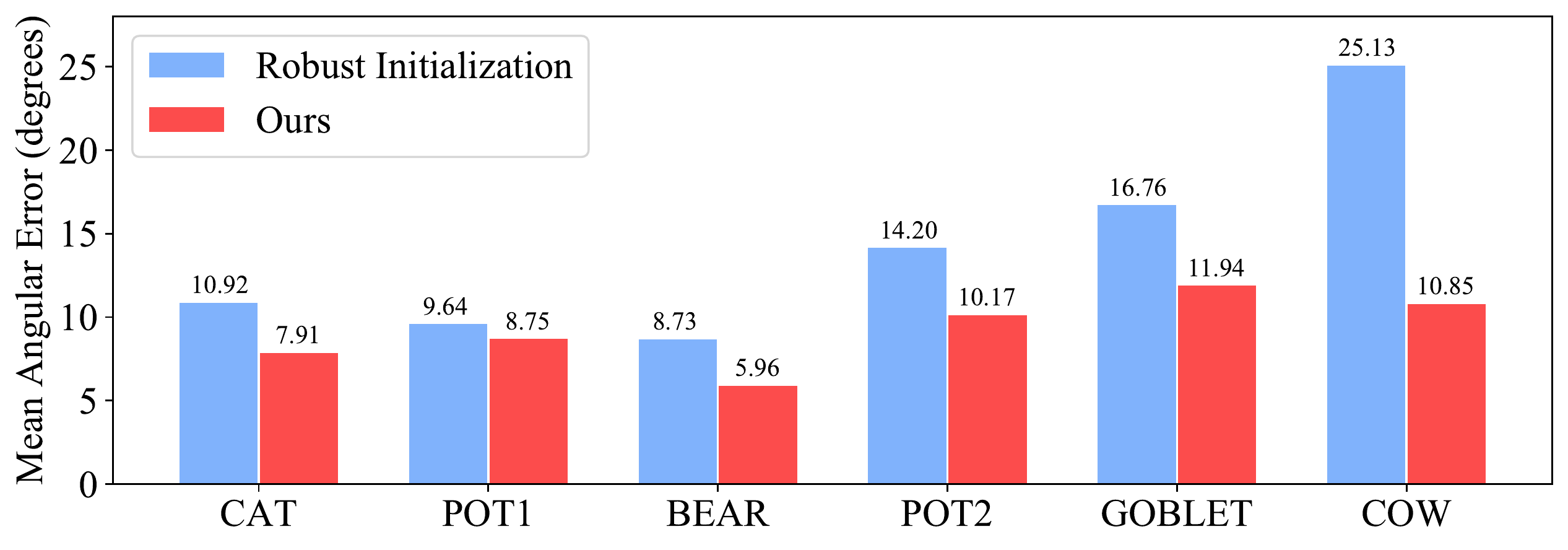}
    \caption{\footnotesize Surface normal accuracy achieved w.r.t its initialization. }
    \label{fig:diligent_bar}
\end{figure}

\begin{figure}
\centering
\includegraphics[{width=1\linewidth}]{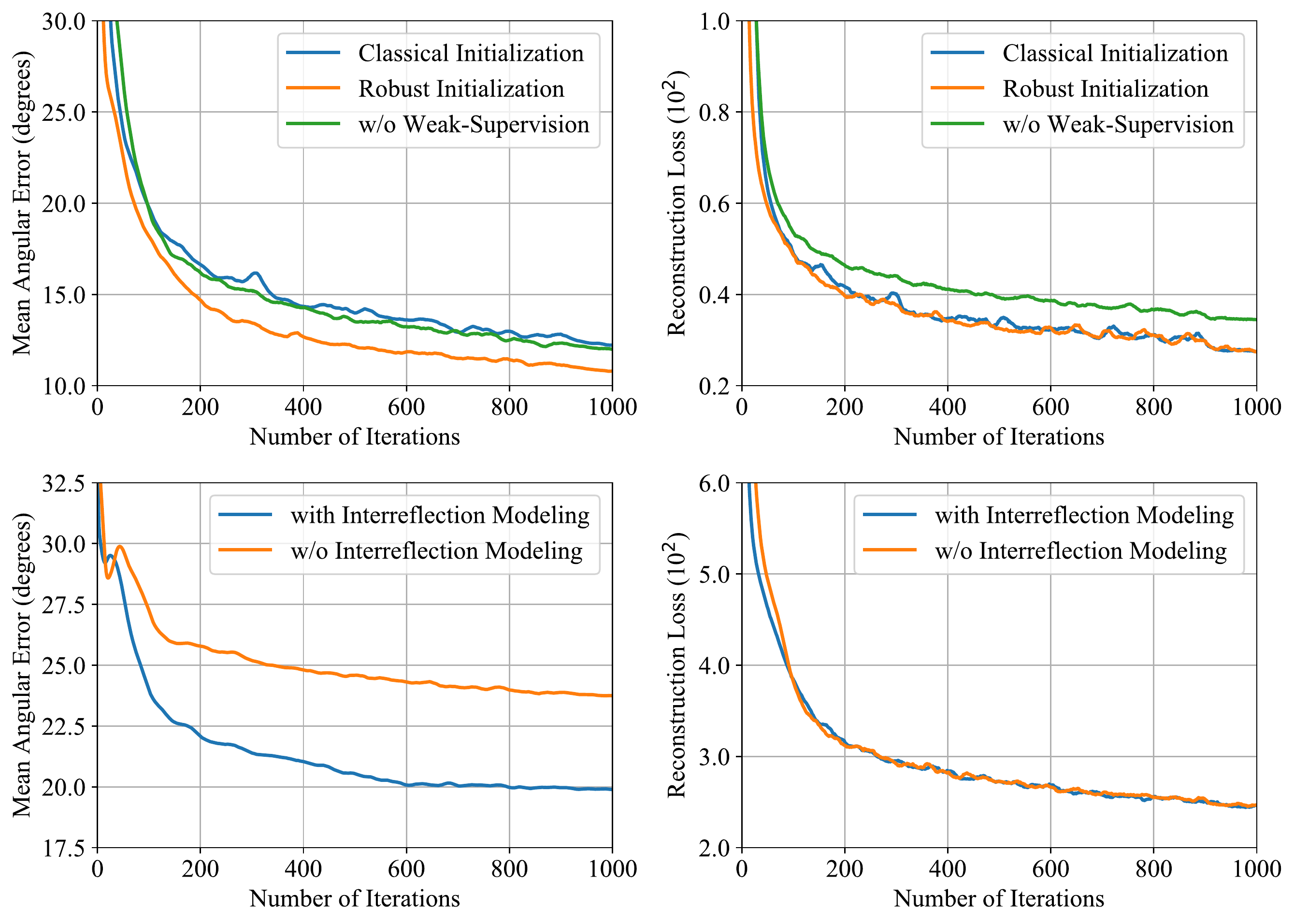}
\caption{\footnotesize \textbf{Ablation Study:}   
We demonstrate the effect of robust initialization on  \textit{Cow} (top) and interreflection modeling on \textit{Vase} (bottom).
}
\label{fig:ablation}
\end{figure}

\noindent
\textit{$\bullet$ \uline{Robust Initialization}:}
To show the effect of initialization, we consider three cases. First, we use classical approach \cite{woodham1980photometric} to initialize inverse rendering network. Second, we replace the classical method with our robust initialization strategy. In the final case, we remove the weak-supervision loss from our method. Fig.\ref{fig:ablation} shows MAE and image reconstruction loss curve per learning iteration obtained on \textit{Cow} dataset. The results indicate that robust initialization allows the network to converge faster as outliers are separated from the images at an initial stage. Fig.\ref{fig:diligent_bar} shows the MAE of surface normals during initialization as compared to the results obtained using our method.

\noindent
\textit{$\bullet$ \uline{Interreflection Modeling}:}
To demonstrate the effect of interreflection modeling, we remove the function $\xi_{n2}$ in Eq:\eqref{eq:nayarupdateblock} and use $\mathbf{N}_o$ in image reconstruction as in classical rendering. Fig.\ref{fig:ablation} provides learning curves with and without interreflection modeling. As expected, excluding the effect of interreflections inherently impacts the accuracy of the surface normals estimates even if the image reconstruction quality remains consistent. Hence, it is important to explicitly constrain the geometry information.

\smallskip
\formattedparagraph{(d) Limitations.}
Discrete facets assumption of a continuous surface for computing depth and interreflection kernel may not be suitable where the surface is discontinuous in orientation, \eg, surface with deep holes, concentric rings, \etc. As a result, our method may fail on surfaces with very deep concavities and cases related to naturally occurring optical caustics. As a second limitation, the light estimation network may not resolve GBR ambiguity for all kinds of shapes. Presently, we did not witness such ambiguity with the light calibration network as it is trained to predict lights under non-GBR transformed surface material distribution.


\section{Conclusion}
From this work, we conclude that uncalibrated neural inverse rendering approach with explicit interreflection modeling enforces the network to model complex reflectance characteristics of objects with different material and geometry types. Without using ground-truth surface normals, we observed that our method could provide comparable or better results than the supervised approaches. And therefore, our work can enable 3D vision practitioners to opt for photometric stereo methods to study a broader range of geometric surfaces. That's said,  image formation is a complex process, and additional explicit constraints based on the 3D surface geometry types, material, and light interaction behavior could further advance our work.

\smallskip
\formattedparagraph{Acknowledgement.} {{This work was funded by Focused Research Award from Google (CVL, ETH 2019-HE-318, 2019-HE-323). We thank Vincent Vanweddingen from KU Lueven for providing some datasets for our experiments.}}

{\small
\bibliographystyle{ieee_fullname}
\bibliography{cvpr_camera_ready}
\nocite{wang2020non}
\nocite{yao2020gps}
\nocite{Asselin_2020}
\nocite{chen2019microfacet}
\nocite{zheng2019spline}
\nocite{li2019learning}
\nocite{wang2020lightweight}
\nocite{enomoto2020photometric}
\nocite{logothetis2019differential}
\nocite{li2020multi}
\nocite{park2013multiview}
\nocite{sengupta2018sfsnet}
\nocite{bi2020deep}
\nocite{blinn1977models}
\nocite{hale2008fixed}
\nocite{johnson2011shape}
\nocite{Wiles_2017}
\nocite{matusik2003data}
}

\appendix
\twocolumn[\section*{\Large [Supplementary Material] Uncalibrated Neural Inverse Rendering for Photometric Stereo of General Surfaces}]

\begin{abstract}
In our supplementary material, we first present a few case studies to analyze our method's effectiveness. Next, we give a detailed description of our coding implementation for training and testing the neural network outlined in the main paper. Formally, this report includes the coding platform details ---both hardware and software, with train and test time observed across different datasets. Further, mathematical derivations of our robust initialization and specular-reflectance map formulations are supplied. Finally, we analyze the light estimation performance and discuss the possible future extensions of our method. Besides, our supplementary material includes a short video clip that illustrates the image acquisition setup and visual results.
\end{abstract}


\section{Case Study}
This section provides the observation on the case study that we conducted for our proposed method. It is done to analyze the behavior of our method under different possible variations in our experimental setup. Such a study can help us understand the behavior, pros, and cons of our approach.

\formattedparagraph{Case Study 1:} \emph{What if we use ground-truth light as input to inverse rendering network instead of relying on light estimation network?}

\noindent

This case study investigates the reliability of our method. To conduct this experiment, we supplied ground-truth light source directions and intensities as input to the inverse rendering network and robust initialization. The goal is to study the expected deviation in the accuracy of surface normals when ground-truth light sources information is used, compared to the light calibration network. Table (\ref{tab:calibratedmethods}) compares our method's performance with recent deep calibrated photometric stereo methods on our proposed dataset. The results show that our inverse rendering method achieves the best performance in the calibrated setting, although it does not use a training dataset like other deep-learning-based methods.  Additionally, we observed that the CNNPS model proposed by Ikehata \cite{ikehata2018cnn} which performs per-pixel estimation using observation maps, may not provide accurate surface normals for interreflecting surfaces such as the \textit{Vase} and the \textit{Broken Pot}. Hence, we conclude that extracting information by utilizing the surface geometry is crucial for solving photometric stereo since all surface points affect each other.

Moreover, in Table (\ref{tab:calibratedmethods}), we show the comparison of our method's performance under calibrated and uncalibrated settings. Our method achieves $12.68\degree$ MAE on average, using ground-truth light as input. At the same time, it reaches an average MAE of $14.74\degree$ utilizing the information of the light source obtained from the light estimation network. The difference between these two scores is $2.06$ degrees, which indicates that the gap between the calibrated and uncalibrated settings is not substantial. Accordingly, we can conclude that our method is robust to the variations in the estimated lighting. Further, we observed that our method performs better with the network estimated light sources information in the categories like \textit{Golf-ball}, \textit{Face}. Hence, based on that observation, we can conclude that the availability of ground-truth calibration data is not a strict requirement for achieving better surface normals estimates in photometric stereo for all kinds of surface geometry.

\begin{table*}[t]
\scriptsize
\centering
{
{\begin{tabular}{ c | c | r *{6}{|c} | c  }
\hline
\rowcolor[gray]{0.85}
\textbf{Type} &  \textbf{G.T. Normal} & \textbf{Methods}$\downarrow$ $|$ \textbf{Dataset} $\rightarrow$  & \textbf{Vase} & \textbf{Golf-ball}  &  \textbf{Face}   &  \textbf{Tablet 1} & \textbf{Tablet 2} & \textbf{Broken Pot} &   \textbf{Average Performance} \\
\hline
NN-based & \cmark & Ikehata (2018)\cite{ikehata2018cnn} & 34.00 &
14.96 &
16.61 &
16.64 &
12.32 &
18.31 &
18.81 \\
\hline
NN-based & \cmark &   Chen et al.(PS-FCN)(2018)\cite{chen2018ps} &  
27.11 & 15.99 & 
16.17 &
\textbf{10.23} &
5.79 &
\textbf{8.68} &
14.00 \\
\hline
NN-based & \xmark & Ours (Ground-truth light/ calibrated) & 
\textbf{16.40} & 
14.23 & 
14.24 &
10.77 &
\textbf{4.49} &
15.92 &
\textbf{12.68}  \\
\hline
NN-based & \xmark & Ours (Estimated light/ uncalibrated) & 
19.91 &
\textbf{11.04} &
\textbf{13.43} &
12.37 & 
13.12 &
18.55 &
14.74  \\

\hline
 &  & Diff. in MAE (Ours(Est)-Ours(GT)) &  
+3.51 &
-3.19 &
-0.81 &
+1.60 &
+8.63 &
+2.63 &
+2.06 \\
\hline
\end{tabular}}}
\caption{ \footnotesize Comparison of recent deep \textbf{calibrated} photometric stereo methods Ikehata \cite{ikehata2018cnn} and Chen \etal \cite{chen2018ps} (PS-FCN)  against our method under \textbf{uncalibrated} and \textbf{calibrated} setting. For testing our method under the calibrated setting, we evaluate the performances assuming that ground-truth light source directions and intensities are available. Note that Chen \etal \cite{chen2018ps} and Ikehata \cite{ikehata2018cnn} additionally uses ground-truth surface normals for training, in contrast to our method. The last row shows the difference between our method results when used under uncalibrated and calibrated setting respectively. We can see that the average difference in MAE between the two settings of our method is not significant.  
}
\label{tab:calibratedmethods}
\end{table*}

\formattedparagraph{Case Study 2:} \emph{What if we use noisy images?}

\begin{figure}
    \centering
    \includegraphics[width=1.0\linewidth]{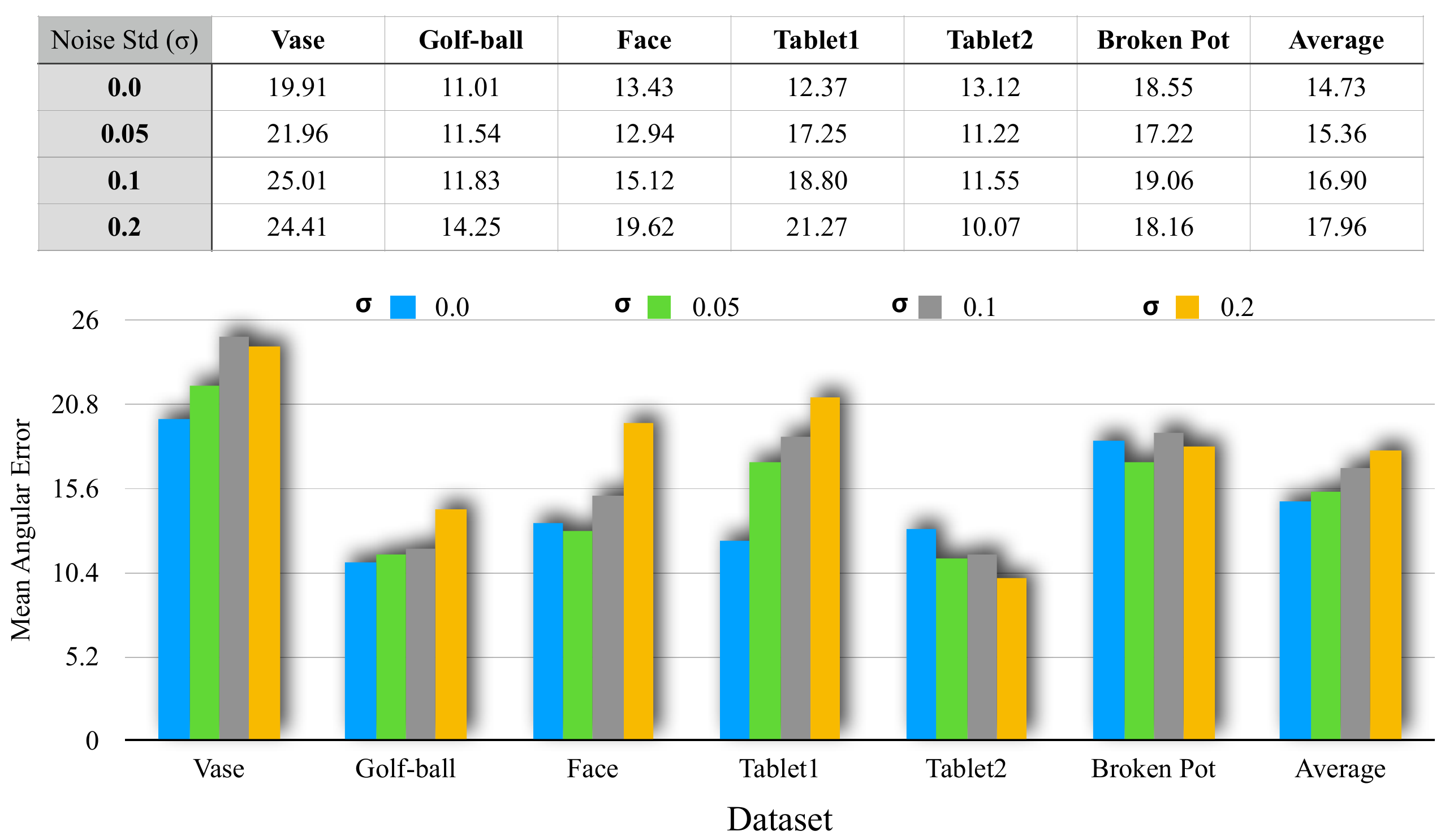}
    \caption{\footnotesize The performance of our method against different noise levels. We used zero-mean Gaussian noise ($\mu$ = 0) with different standard deviations ($\sigma$). We observed that increasing the noise level generally degrades the performance. Still, the behavior under different noise levels varies among the subjects as the performance depends on the signal-to-noise ratio of the images. 
     }
    \label{fig:noise_experiment}
\end{figure}

Photometric stereo uses a camera acquisition setup, and this implies that noise due to imaging is inevitable. 
This case study aims to investigate the behavior of our method on different noise levels. To study such a behavior, we synthesized images by adding noise to the images of our proposed dataset. Fig.\ref{fig:noise_experiment} compares the performance of our method under different noise levels. For this case study, we used zero-mean Gaussian noise with different standard deviations ($\sigma$=0.05, $\sigma$=0.1, $\sigma$=0.2). The quantitative results indicate that increasing the noise generally degrades the performance. We observed that the behavior under different noise levels varies among the subjects. 

\begin{figure}
\centering
    \includegraphics[{width=\linewidth}]{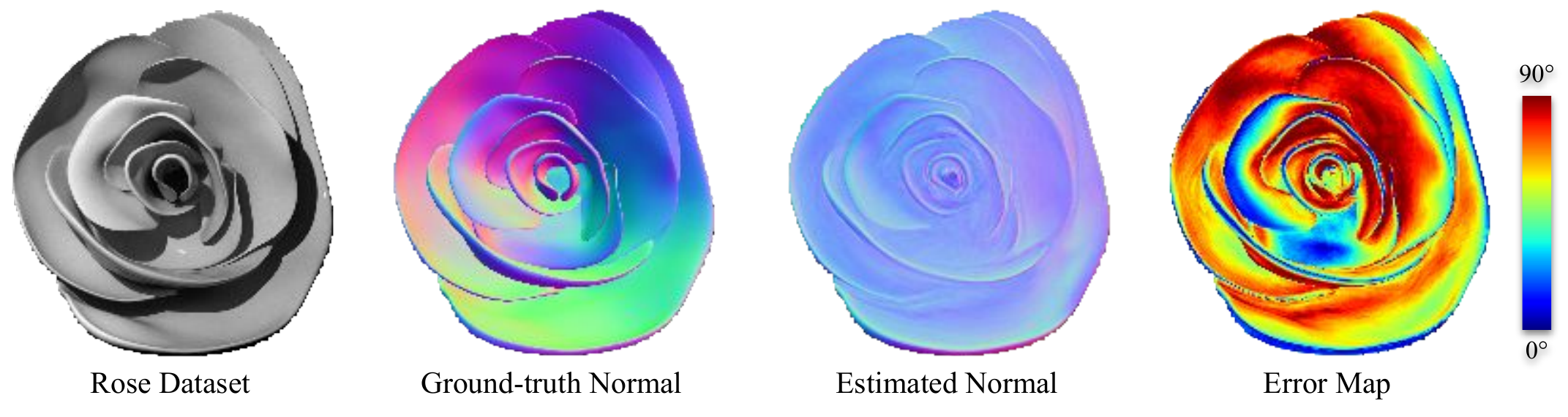}
\caption{ \textbf{Failure case:} Qualitative results on the Rose dataset.
}
\label{fig:Rose}
\end{figure}

\formattedparagraph{Case Study 3:} \emph{Photometric stereo on concentric surfaces with deep concavities and large surface discontinuity.}

To study our photometric stereo method's boundary-condition, we took a complex geometric structure with concentric surfaces, deep-concavities, and large discontinuities for investigation. Accordingly, we synthesized the \emph{Rose} dataset using the same dome-settings outlined in the main paper. Fig.\ref{fig:Rose} shows the qualitative results obtained on this dataset. Our method achieves $60.82$ degrees of MAE on this particular example. We observed that our approach could not handle this complex geometry because the surface is highly discontinuous with excessive gaps between the leaves. The scene is also affected by occlusions and cast shadows, and therefore, modeling the interreflections for this case seems very difficult.

Though our method applies to a broad range of objects, our interreflection modeling is inspired by Nayar \etal \cite{nayar1991shape} formulation, which may not hold for all kinds of surfaces. The interreflection modeling computes depth from the normal map under the continuous surface assumption, which fails in this case study. Furthermore, it models continuous surfaces with discrete facets. Due to such limitations, our method may not be suitable for concentric surfaces with deep concavities and large discontinuities. In such cases, the interreflection effect is very complicated, and our approach may disappoint to model such complex light phenomena.

\section{Coding Details}
This section provides a detailed description of our source code implementation. We start by introducing the light estimation network's training phase. Then we focus on the testing phase, where the inverse rendering network is optimized to estimate the surface normals, depth, and BRDF values. Finally, we present details on training and testing run-times. 

\subsection{Training Details}
As our inverse rendering network optimizes its learnable parameters at the test time, we apply a training stage only to the light estimation network. 
For training the network, we used Blobby and Sculpture datasets that are introduced by Chen \etal \cite{chen2018ps}. This dataset is created by using 3D geometries of Blobby \cite{johnson2011shape}, and Sculpture \cite{Wiles_2017} shape datasets and combining them with different material BRDFs taken from MERL dataset \cite{matusik2003data}. In total, the complete dataset contains $85212$ subjects. For each subject, there exist 64 renderings with different light source directions. The intensity of the light sources is kept constant during the whole data generation process. To simulate different intensities during training, image intensity values are randomly generated in the range of $[0.2, 2]$, and these intensity values are used to scale the image data linearly. In each training iteration, the input data is perturbed in the range of $[-0.025, 0.025 ]$ for augmentation.

\begin{figure*}[t]
\centering
    \includegraphics[{width=0.90\linewidth}]{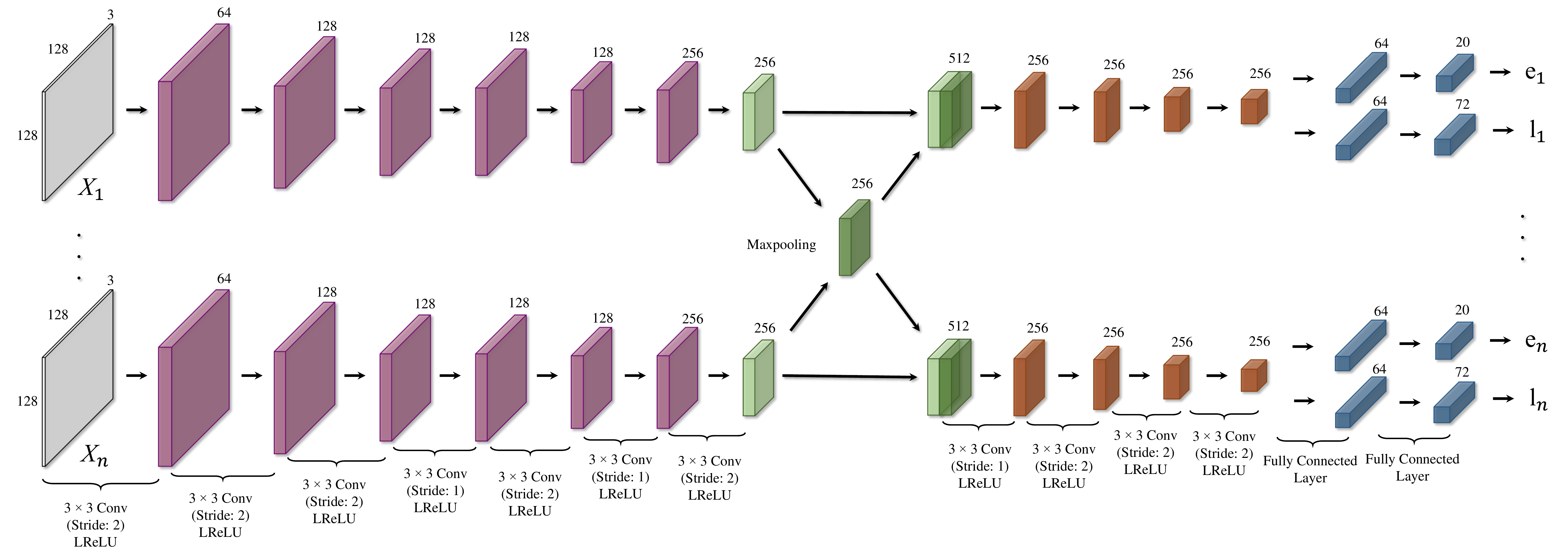}
\caption{ \textbf{Architecture of the light estimation network.} \footnotesize The network first extracts features from the input images separately using feature extraction layers (purple). Then, the extracted image-specific features (light-green) are fused with max-pooling operation to obtain a global representation of the entire scene (dark-green). Finally, all image-specific and global features are used in classifier network where convolution (brown) and fully-connected (blue) layers are used to predict light intensity values  ($e_i$'s) and direction vectors  ($\mathbf{l}_i$'s).
}
\label{fig:LCNET_diagram}
\end{figure*}

The light estimation network is a multiple-input multiple-output (MIMO) system which requires images of the same object captured under different illumination conditions (see Fig.\ref{fig:LCNET_diagram}). The core idea is that all input images have the same surface, and having more images helps the network extract better global features. During training, we use 32 images of the same object for global feature extraction. Note that all of the images are used for feature extraction at test time to achieve the best performance from the network.

\subsection{Testing Details}

Given a set of test images $\mathbf{X}$ and object mask $\mathbf{O}$, we first use the light estimation network to have light source directions and intensities. However, the light estimation network operates on $128\times128$ images because it uses fully connected layers for classification, and these layers process only fixed-length vectors. Consequently, we scale the input images into the resolution of $128\times128$ before feeding them to the network. We apply this pre-processing step only for the light estimation network and use the original image size for all other operations during testing.

Once we obtain the light source directions and intensities, we apply the robust initialization algorithm to get an initial surface normal matrix $\mathbf{N}_{init}$. It also provides an albedo map that is transformed into $\mathbf{P} \in \mathbb{R}^{m \times m} $ which is required for interreflection modeling. Details about the robust initialization method are explained and derived in \S\ref{sec:robust_initialization}. 

After the robust initialization process, we start the optimization of our inverse rendering framework. First, we initialize all the network parameters ($\Theta_f$, $\Theta_{n1}$, $\Theta_{sp}$, $\Theta_{lg}$ $\Theta_{ri}$) which correspond to the weights of the convolution operations.
In this step, we initialize the weights randomly by sampling from a Gaussian distribution with zero mean and $0.02$ variance.  We perform 1000 iterations in total using Adam optimizer \cite{DBLP:journals/corr/KingmaB14} with an initial learning rate of $8 \times 10^{-4}$.  The learning rate is reduced by a factor of 10 after 900 iterations for fine-tuning. We observed that setting these hyperparameters may result in convergence problems in our dataset. For this reason, we set the initial learning rate of the estimation branch ($\xi_{f}$ and $\xi_{n1}$) to $8 \times 10^{-5}$ while experimenting on our dataset. We also inject Gaussian noise with zero mean and $0.1$ variance to the images before feeding them to $f_{sp}$ for image reconstruction. We observed that this prohibits the network from generating degenerate solutions. At every 100 iterations, we update the depth and the interreflection kernel matrix entries using the normal estimation $\mathbf{N}_o$.

\noindent
\textbf{(a) Depth}: To compute the depth from normals, we use a gradient-based method with surface orientation constraint \cite{antensteiner2018review}. Given the surface normals, we first compute a gradient field $\hat{\mathbf{G}} \in \mathbb{R}^{h \times w \times 2}$ where $h$ and $w$ are the spatial dimensions. The idea is that the gradient field computed from surface normal map and the estimated depth $\mathbf{D} \in \mathbb{R}^{h \times w}$ should be consistent, \ie, $\nabla \mathbf{D} \approx \hat{\mathbf{G}}$. That corresponds to an overdetermined system of linear equations and is solved by minimizing the following objective function \ie, Eq:\eqref{eq:solutiondepth} using the least-squares approach

\begin{equation}\label{eq:solutiondepth}
    \begin{aligned}
        & \displaystyle \underset{\mathbf{D}} {\textrm{min.}} ~\|\nabla \mathbf{D} - \hat{\mathbf{G}}\|^2
    \end{aligned}
\end{equation}

\begin{figure*}
\centering
    \includegraphics[{width=1\linewidth}]{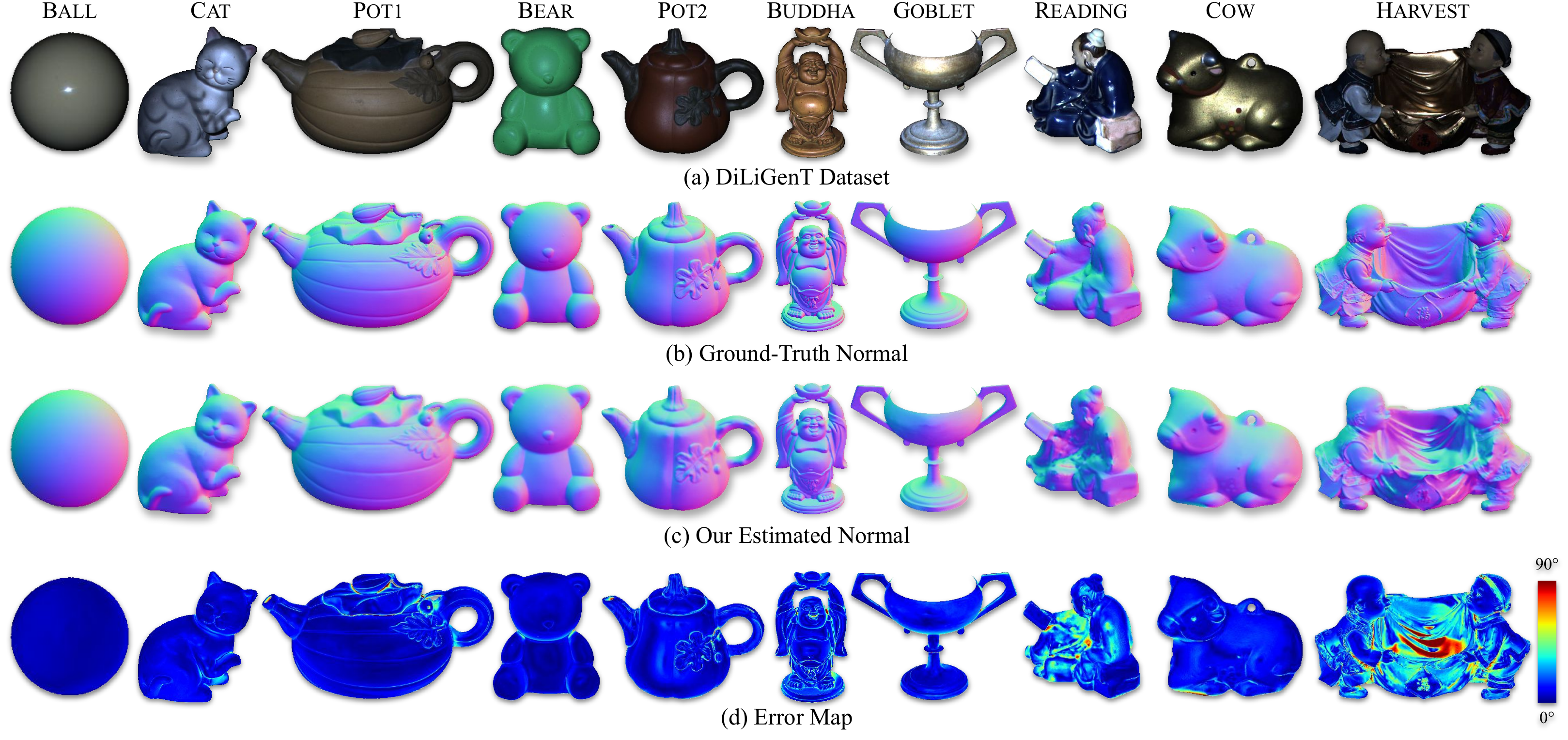}
\caption{We present visual results of our method on all of the DiLiGenT categories. The bottom row demonstrates the angular error maps obtained form our estimations and ground-truth normals. 
}
\label{fig:all_diligent_images}
\end{figure*}

\noindent
\textbf{(b) Interreflection Modeling}: To consider the effect of interreflection during the image reconstruction process, we define the function $\xi_{n2}$ which uses the estimated normal $\mathbf{N}_o  \in \mathbb{R}^{3 \times m} $, albedo matrix $\mathbf{P} \in \mathbb{R}^{m \times m}$ and the interreflection kernel $\mathbf{K} \in \mathbb{R}^{m \times m}$. Given all these components, Nayar \etal \cite{nayar1991shape} relates the observed radiance ($\mathbf{X}$) and the radiance due to primary light source ($\mathbf{X}_\mathbf{s}$) as follows:

\begin{equation}
    \begin{aligned}\label{eq:labelnayerupdate}
        \mathbf{X} = (\mathbf{I} - \mathbf{P}\mathbf{K})^{-1}\mathbf{X}_\mathbf{s}
    \end{aligned}
\end{equation}

Assuming the surface shows Lambertian reflectance property, we model the radiance in terms of facet matrices as follows:

\begin{equation}
    \begin{aligned}\label{eq:facetupdate}
        \mathbf{X} = \mathbf{F}_{ny} \mathbf{L},
        ~~ \mathbf{X}_\mathbf{s} = \mathbf{F} \mathbf{L},
        ~~\Rightarrow 
        \mathbf{F}_\mathbf{ny} = (\mathbf{I} - \mathbf{P}\mathbf{K})^{-1}\mathbf{F}
    \end{aligned}
\end{equation}

Here $\mathbf{F}_{ny} \in \mathbb{R}^{m \times 3}$ and $\mathbf{F} \in \mathbb{R}^{m \times 3}$ are the facet matrices which contain surface normals $\mathbf{N}_{ny}$ and $\mathbf{N}_o$ scaled with local reflectance value. We use Eq:\eqref{eq:facetupdate} to obtain $\mathbf{F}_{ny}$ and normalize each row to unit vector to obtain $\mathbf{N}_{ny}$. 

The computation of the interreflection kernel $\mathbf{K}$ has the complexity of $\mathcal{O}(n^2)$ where $n$ is the number of facets. Therefore, treating each pixel as a facet limits the application of our method. To approximate the effect of interreflections, we downsample the normal maps with the factor of $4$ and calculated the kernel values accordingly. After the normal is updated, we scale it to the original size managing the image details appropriately.

\subsection{Timing Details}

Our framework is implemented in Python using PyTorch version $1.1.0$. Table (\ref{tab:timing_details}) provides the light estimation network's training time and the inference time of neural inverse rendering network on two datasets separately.

\section{Mathematical Derivations}
Here, we supply the mathematical derivation pertaining to the initialization of the surface normals to the inverse rendering network. For completion,  we also supplied the well-known deviation of reflection vector \S \ref{ssec:specular_ref}.

\subsection{Robust Initialization} \label{sec:robust_initialization}
Our surface normals initialization procedure aims at recovering the low rank matrix $\mathbf{Z}\in \mathbb{R}^{m \times n}$ from the image matrix $\mathbf{X} \in \mathbb{R}^{m \times n}$ such that $\mathbf{X} = \mathbf{Z} + \mathbf{E} $ where $\mathbf{E} \in \mathbb{R}^{m \times n}$ is the matrix of outliers. Here, we assume that the low-rank matrix follows the classical photometric stereo model  ($\mathbf{Z} = \mathbf{N}^{\mathrm{T}}\mathbf{L}$) and the outlier matrix $\mathbf{E}$ is sparse in its distribution. Since it is known by definition that $\mathbf{Z}$ spans a rank-3 space, it can be formulated as a standard RPCA problem \cite{wu2010robust}. However, we know that RPCA formulation performs the nuclear norm minimization of $\mathbf{Z}$ matrix which not only minimizes the rank but also the variance of $\mathbf{Z}$ within the target rank. Now, for the photometric stereo model, it is easy to infer that $\mathbf{N}$ lies in a rank 3 space.  As the true rank for $\mathbf{Z}$ is known from its mathematical construction, we do not want to minimize the subspace variance within the target range. Nevertheless, this strict constraint is difficult to meet due to the complex imaging model, and therefore, we encourage to preserve the variance of information within the target range while minimizing the other singular values outside the target rank ($K$). So, we minimize the partial sum of the singular values which are outside the target rank with the following  optimization as follows:

\begin{equation}\label{eq:pami15ps1}
\begin{aligned}
\centering
& \displaystyle \underset{\mathbf{Z}, \mathbf{E}} {\textrm{minimize}} ~\|\mathbf{Z}\|_{r=K} + \lambda\|\mathbf{E}\|_1, ~~\textrm{subject to:}~\mathbf{X} = \mathbf{Z} + \mathbf{E} 
\end{aligned}
\end{equation}

\begin{table}
\centering
\resizebox{\columnwidth}{!}
{\begin{tabular}{|r|c|c|}
\hline
\rowcolor[gray]{0.85}
&  \textbf{GPU}  &  \textbf{Time}  \\
\hline
Training of Light Estimation Network & Titan X Pascal (12GB) &  $\approx$ 22 hours \\
\hline
Inference on DiLiGenT & GeForce GTX TITAN X (12GB) &  $53.41 \pm 41.57$ min per subject  \\ 
\hline
Inference on our Dataset & GeForce GTX TITAN X (12GB) & $ 29.08 \pm 15.99$ min per subject  \\ 
\hline

\end{tabular}}
\caption{\footnotesize Measured training and testing time with respect to the utilized hardware. For our dataset, we have 100 to 260 images per subject and the DiLiGenT dataset has 96 images per subject. Note: Deep photometric stereo method processes a set of images rather than one image for estimating normals.}
\label{tab:timing_details}
\end{table}

The Augmented Lagrangian function of Eq:\eqref{eq:pami15ps1} can be written as follows:

\begin{equation}\label{eq:augmented_lagrangian}
    \begin{aligned}
        & \displaystyle \mathcal{L}(\mathbf{Z}, \mathbf{E}, \mathbf{Y}) = \|\mathbf{Z}\|_{r=K} + \lambda\|\mathbf{E}\|_1 + \frac{\mu}{2}\|\mathbf{X}-\mathbf{Z}-\mathbf{E}\|_{F}^2 + \\
        & \displaystyle  <\mathbf{Y}, \mathbf{X}-\mathbf{Z}-\mathbf{E}>
    \end{aligned}
\end{equation}

Here, $\mu$ is a positive scalar and $\mathbf{Y} \in \mathbb{R}^{m \times n}$ is the estimate of the Lagrange multiplier. As minimizing this function is challenging, we solve it by utilizing the alternating direction method of multipliers (ADMM)\cite{boyd2011distributed, oh2013partial, lin2010augmented}. Accordingly, the optimization problem in Eq:\eqref{eq:augmented_lagrangian} can be divided into sub-problems, where $\mathbf{Z}$, $\mathbf{E}$ and $\mathbf{Y}$ are updated alternatively while keeping the other variables fixed.

\noindent
\textbf{1. Solution to Z:}
\begin{equation}\label{eq:solutionZ}
    \begin{aligned}
        & \displaystyle \mathbf{Z}^{*} = \displaystyle \underset{\mathbf{Z}} {\textrm{argmin}} ~\|\mathbf{Z}\|_{r=K} + \frac{\mu_{k}}{2}\|\mathbf{Z} - (\mathbf{X}-\mathbf{E}_{k} + \mu_{k}^{-1}\mathbf{Y}_k)\|_{F}^2
    \end{aligned}
\end{equation}

The solution to Eq:(\ref{eq:solutionZ}) sub-problem  at $k^{th}$ iteration is given by $\mathbf{Z}_{k} = \mathcal{P}_{K, \mu_k^{-1}}[\mathbf{X}-\mathbf{E}_{k} + \mu_{k}^{-1}\mathbf{Y}_k]$ where, $\mathcal{P}_{K, \tau}[\mathbf{M}] = \mathbf{U_M} (\Sigma_{\mathbf{M_1}} + \mathcal{S}_{\tau}[\Sigma_{\mathbf{M_2}}])\mathbf{V}_{\mathbf{M}}^T$ is the partial singular value thresholding operator \cite{oh2013partial} and $\mathcal{S}_{\tau}[x] = \textrm{sign}(x)\max(|x|-\tau, 0)$ is the soft-thresholding operator \cite{hale2008fixed}. Here, $\mathbf{U_M}, \mathbf{V_M}$ are the singular vector of matrix $\mathbf{M}$ and $\Sigma_{\mathbf{M_1}} = \textbf{diag}(\sigma_1, \sigma_2,...\sigma_K, 0, 0)$, $\Sigma_{\mathbf{M_2}} = \textbf{diag}(0, 0,..,\sigma_{K+1},..,\sigma_{N})$.\\

\noindent
\textbf{2. Solution to E:}
\begin{equation}\label{eq:solutionE}
    \begin{aligned}
        & \displaystyle \mathbf{E}^{*} = \displaystyle \underset{\mathbf{E}} {\textrm{argmin}} ~\lambda\|\mathbf{E}\|_1 + \frac{\mu_{k}}{2}\|\mathbf{E} - (\mathbf{X}-\mathbf{Z}_{k+1} + \mu_{k}^{-1}\mathbf{Y}_k)\|_{F}^2
    \end{aligned}
\end{equation}
The solution to Eq:(\ref{eq:solutionE}) sub-problem at $k^{th}$ iteration is given by $\mathbf{E}_{k} = \mathcal{S}_{\lambda\mu_{k}^{-1}}[\mathbf{X}-\mathbf{Z}_{k+1} + \mu_{k}^{-1}\mathbf{Y}_k]$ where, $\mathcal{S}_{\tau}[x] = \textrm{sign}(x)\max(|x|-\tau, 0)$ is a soft-thresholding operator \cite{hale2008fixed}.
For proof of convergence and theoretical analysis of partial singular value thresholding operator kindly refer to Oh \etal \cite{oh2013partial} work.
We solve for $\mathbf{Z}$, $\mathbf{E}$ using ADMM until convergence for $K=3$ and use the obtained surface normals for initializing the loss function of inverse rendering network.\\

\noindent
\textbf{3. Solution to Y:} The variable $\mathbf{Y}$ is updated as follows over the iteration:

\begin{equation}
    \mathbf{Y}_{k+1} = \mathbf{Y}_k + \mu_k(\mathbf{X} - \mathbf{Z}_{k+1} - \mathbf{E}_{k+1})
\end{equation}
For more details on the implementation kindly refer to Oh \etal \cite{oh2013partial} method.

\subsection{Derivation of Specular-Reflection Equation 11 in the Main Paper}\label{ssec:specular_ref}

\begin{figure}
\centering
\includegraphics[{width=0.70\linewidth}]{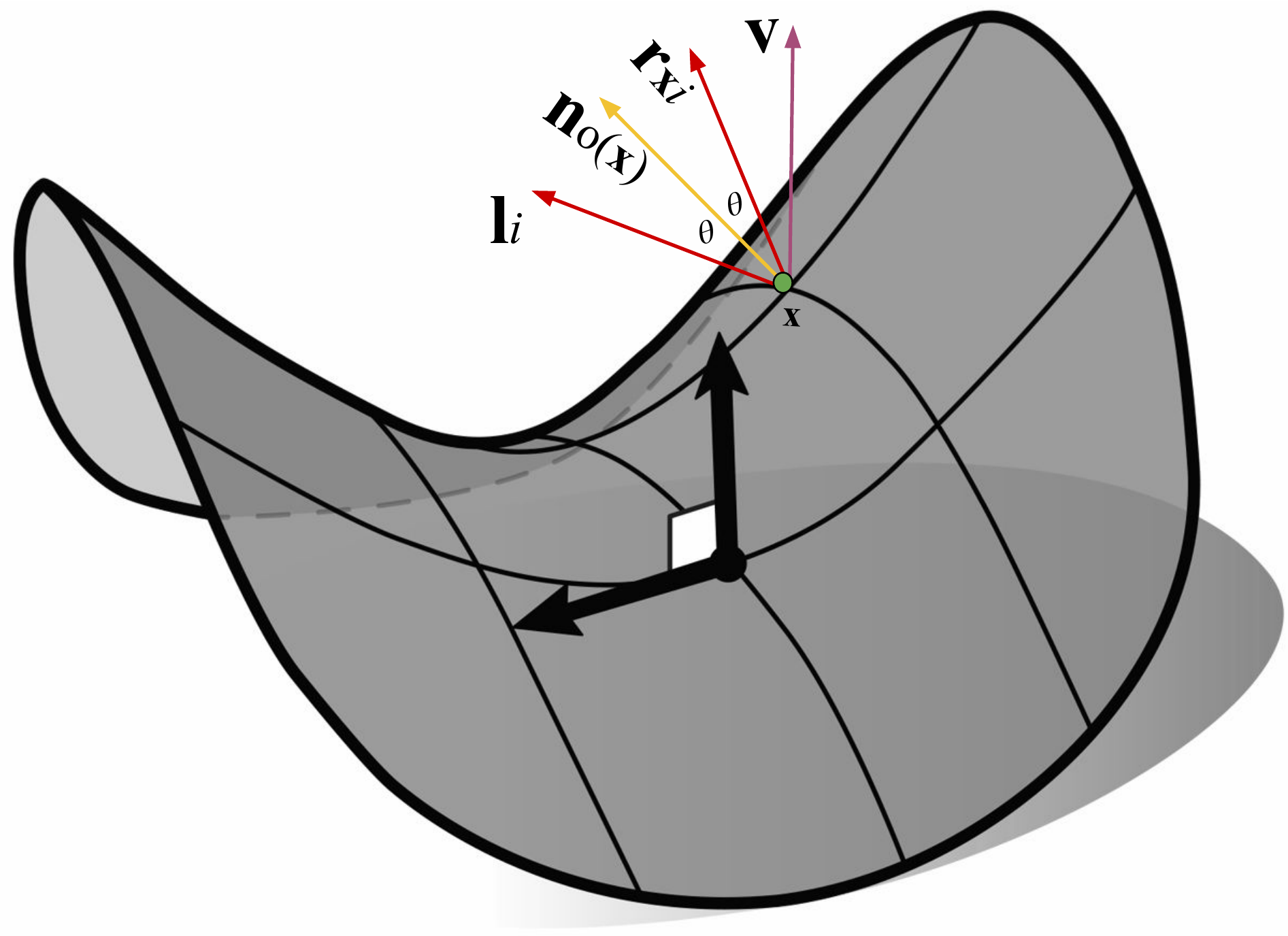}
\caption{Illustration of surface reflectance. When light ray $\mathbf{l}_{i}$ hits a surface element, the specular component along the view-direction of the point x due to $i^{th}$ source is given by $\mathbf{r}_{\mathbf{x}i}$. This presentation of 3D geometry is inspired by Keenan work \cite{Crane:2013:CGP}.
}
\label{fig:specular_reflection_map}
\end{figure}

For completion, we derive Equation 11 of the main paper that is used to compute the specular-reflection map ${R}_i \in \mathbb{R}^{h \times w \times 1}$ for each image. 
 To compute it, we first compute $\mathbf{r}_{ \mathbf{x}i }$ for each point $\mathbf{x}$ that is the direction vector with the highest specular component using the following well-known relation; assuming $\mathbf{l}_i$, and $\mathbf{n}_{o}$ as unit length vectors:

\begin{equation}
    \begin{aligned}
        & \displaystyle \mathbf{r}_{ \mathbf{x}i } + \mathbf{l}_i = 2cos(\theta).\mathbf{n}_{o}(\mathbf{x}); ~~\mathbf{n}_{o}(\mathbf{x})^{T}\mathbf{l}_i = cos(\theta)\\
        & \displaystyle \mathbf{r}_{\mathbf{x}i} = 2\big(\mathbf{n}_{o}(\mathbf{x})^{T}\mathbf{l}_i)\mathbf{n}_{o}(\mathbf{x}\big)-\mathbf{l}_i
    \end{aligned}
\end{equation}
Here, $\mathbf{r}_{ \mathbf{x}i }$ is also a unit length vector (see Fig.\ref{fig:specular_reflection_map}). The component of specular reflection in the view-direction $\mathbf{v} = (0, 0, 1)^{T}$ of the point $\mathbf{x}$ due to $i^{th}$ light is computed as: 
\begin{equation}
    \begin{aligned}
    & \displaystyle \mathbf{r}_{\mathbf{x}i} = \mathbf{v}^{T}\Big(2\big(\mathbf{n}_{o}(\mathbf{x})^{T}\mathbf{l}_i)\mathbf{n}_{o}(\mathbf{x}\big)-\mathbf{l}_i\Big)
    \end{aligned}
\end{equation}
The above relation show that the specular highlights are strongest if the normal $\mathbf{n}_{o}(\mathbf{x})$ is closest to $\mathbf{r}_{ \mathbf{x}i }$. Performing this operation for each point gives us the specular-reflection map $\mathbf{R}_i$.

\begin{figure*}
\centering
\subfigure[\label{fig:calibration_sphere} Calibration Sphere]{\includegraphics[ height=0.19\textwidth]{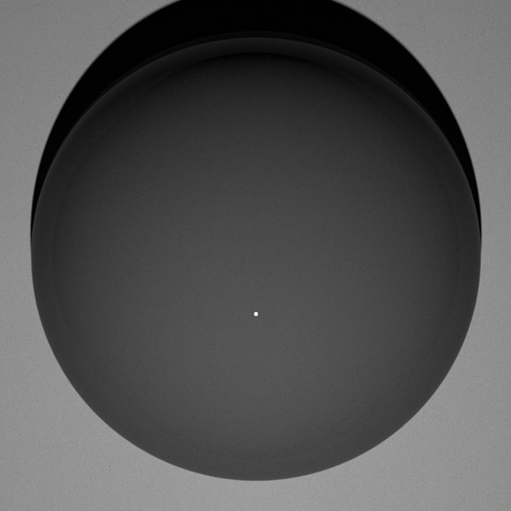}}
\subfigure[\label{fig:x_comp} $x$ Component]{\includegraphics[ height=0.18\textwidth]{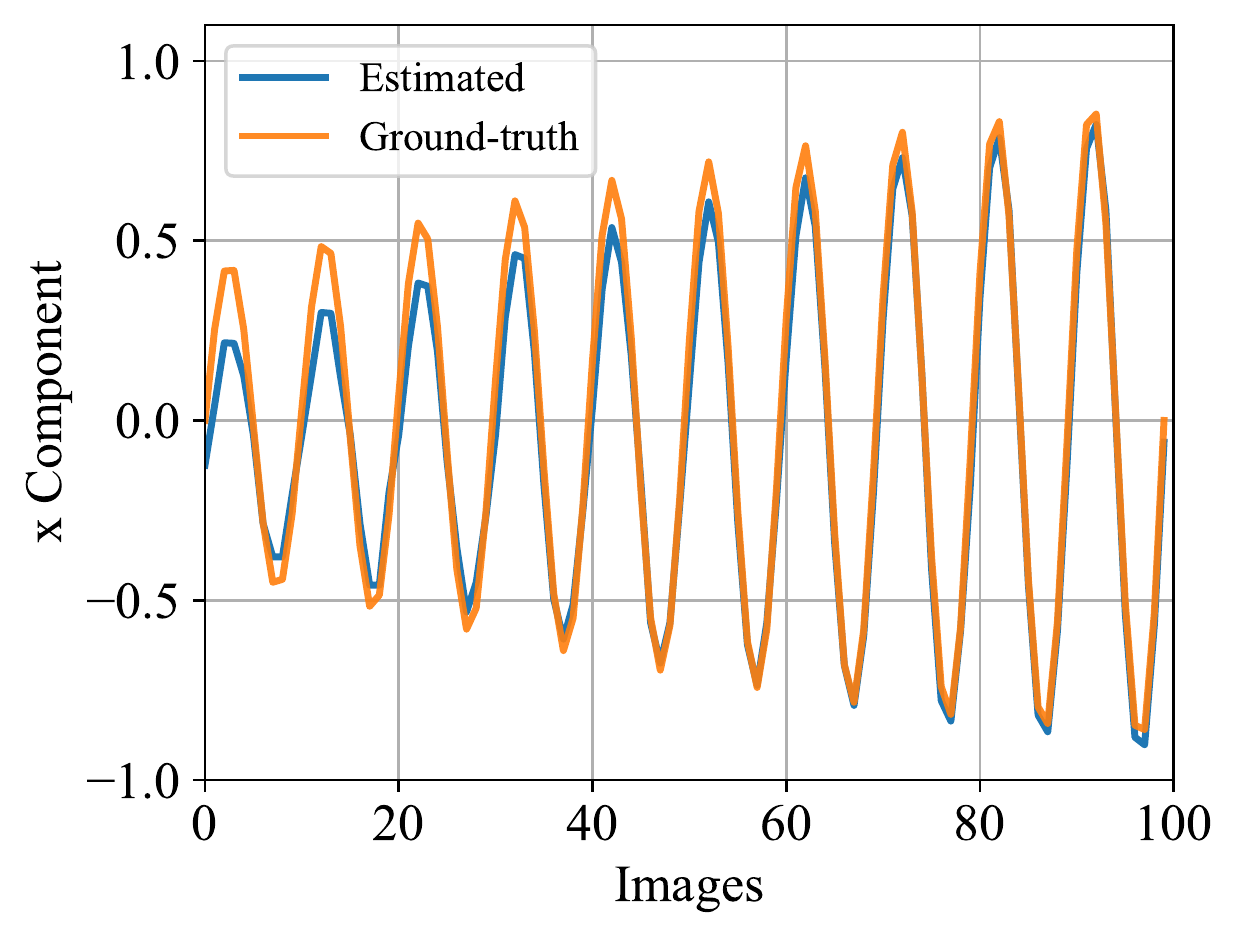}}
\subfigure[\label{fig:y_comp} $y$ Component]{\includegraphics[ height=0.18\textwidth]{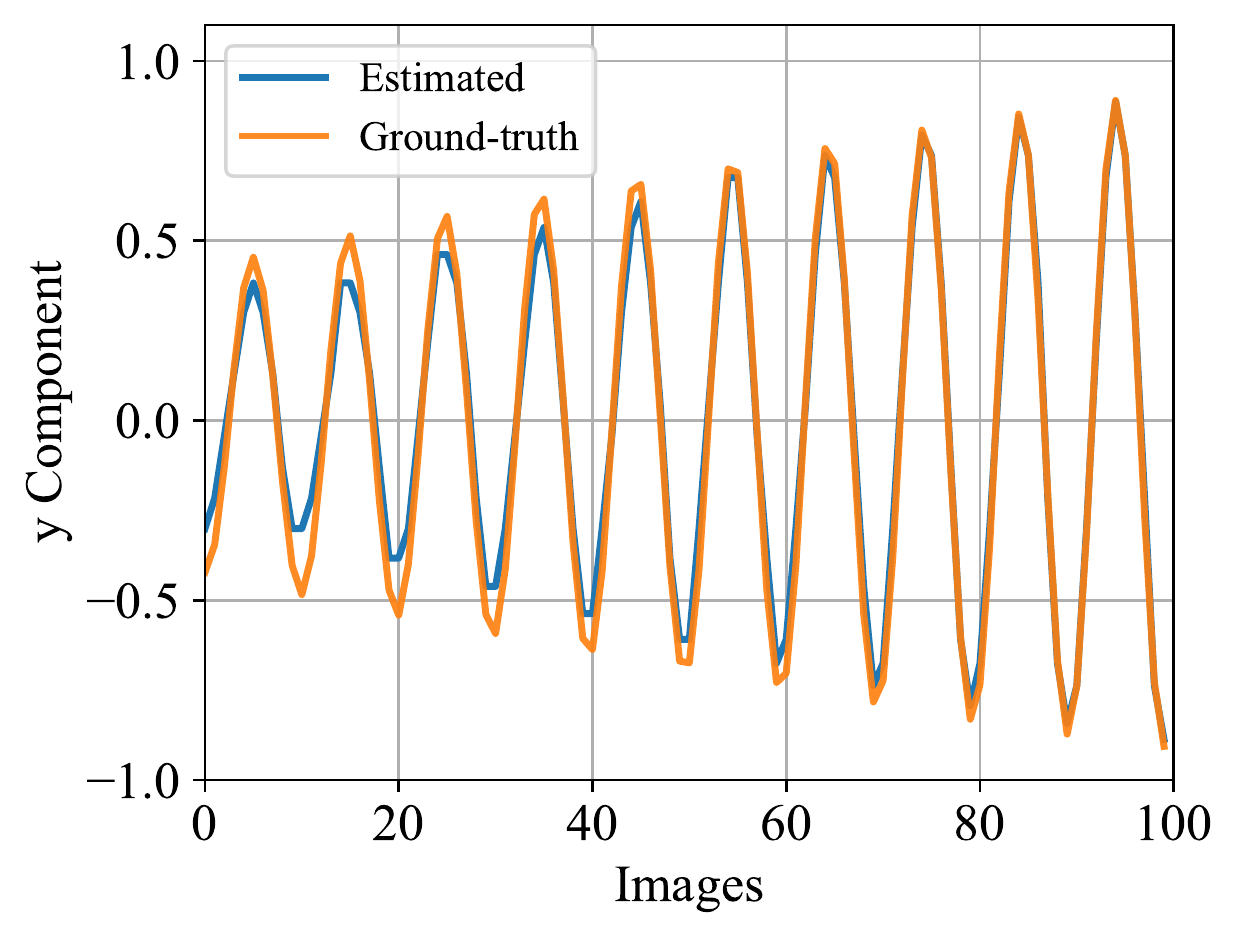}}
\subfigure[\label{fig:z_comp} $z$ Component]{\includegraphics[ height=0.18\textwidth]{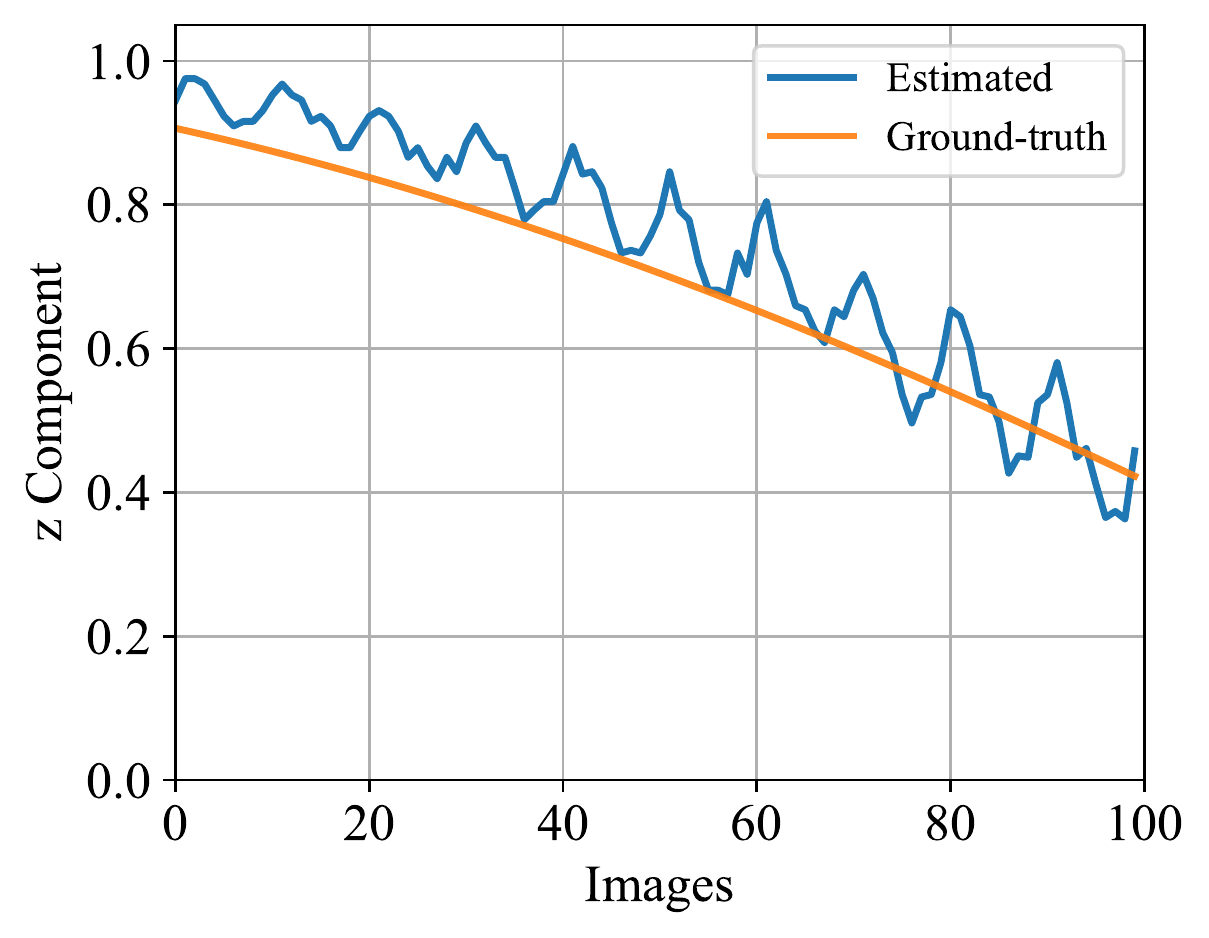}}
\caption{ \footnotesize Light source directions obtained from the calibration sphere (a) using the light estimation network. We demonstrate the $x$,$y$ and $z$ components of the light direction vectors (b-d). The mean angular error between the ground-truth and estimated light directions is $6.31$ degrees. 
}
\label{fig:light_plots}
\end{figure*}

\section{Statistical Analysis of Estimated Light Source Directions}
We aim to investigate the source directions' behavior predicted by the light estimation network (Fig.\ref{fig:LCNET_diagram}). For that purpose, we use a well-known setup used for light calibration, \ie, a calibration sphere.  Our renderings from the calibration sphere (see Fig.\ref{fig:calibration_sphere}) has specular highlights and attached shadows, which provide useful cues for the light estimation network. Figures \ref{fig:x_comp}-\ref{fig:z_comp} illustrate the $x$, $y$ and $z$ components of the estimated light source direction and ground-truth with respect to the images. We measured the MAE between these vectors as $6.31$ degrees. We also observed that the $x$ and $y$ components match well with the ground-truth values.
On the other hand, we observed fluctuations on the $z$ component where the values slightly deviate from the ground-truth in a specific pattern. One possible explanation for this observation is that the network has a bias such that its behavior changes in the different regions of the lighting space. Since we generated the data by moving the light source on a circular pattern around $z$-axis, Fig. \ref{fig:z_comp} also follows a similar pattern with the same frequency with $x$ and $y$ components' curves.

\section{More Qualitative Results Comparison on our Dataset}

\begin{figure}
\centering
    \includegraphics[{width=\linewidth}]{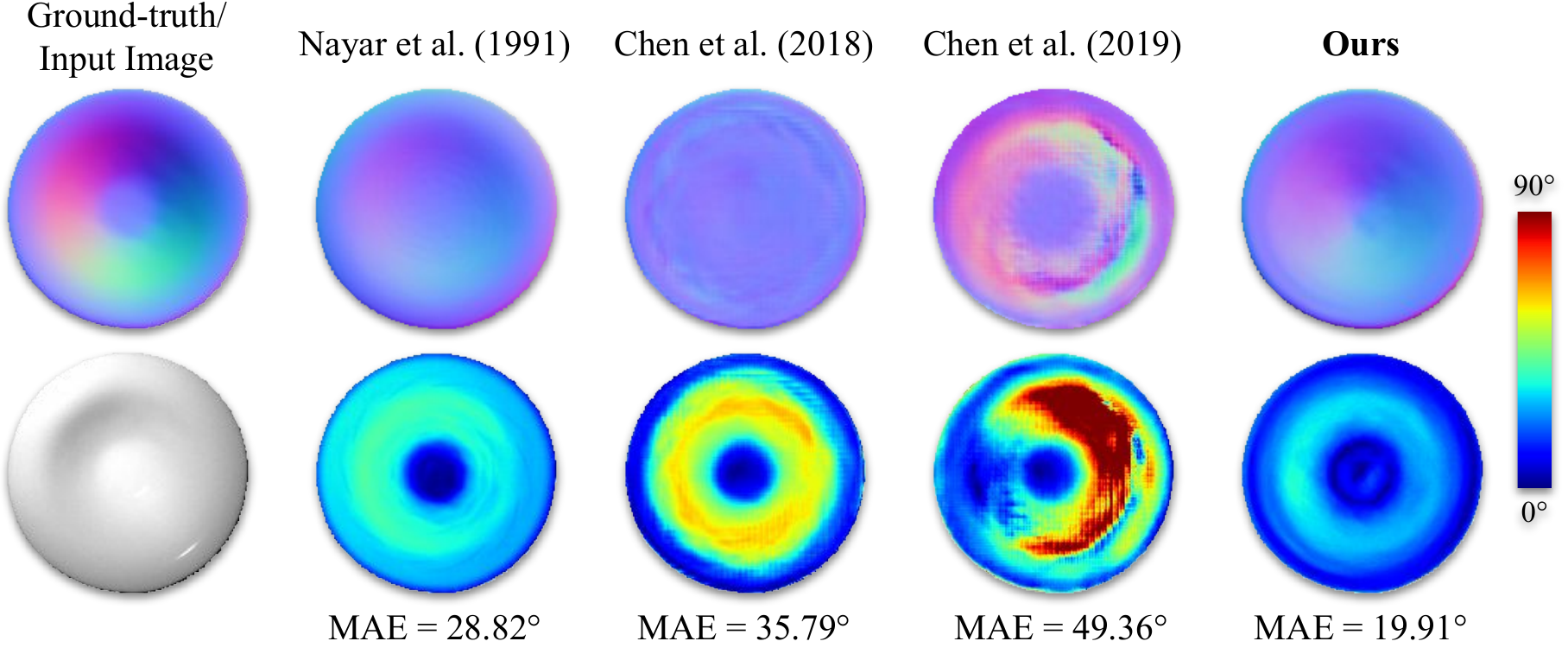}
\caption{ \footnotesize Qualitative comparison on the \textbf{Vase} scene. Here, it is obvious that previous deep learning based methods fail to handle the concavity of the subject. In contrast, our method works reasonably well showing the competence of our modeling procedure.
}
\label{fig:vase}
\end{figure}

\begin{figure}
\centering
    \includegraphics[{width=\linewidth}]{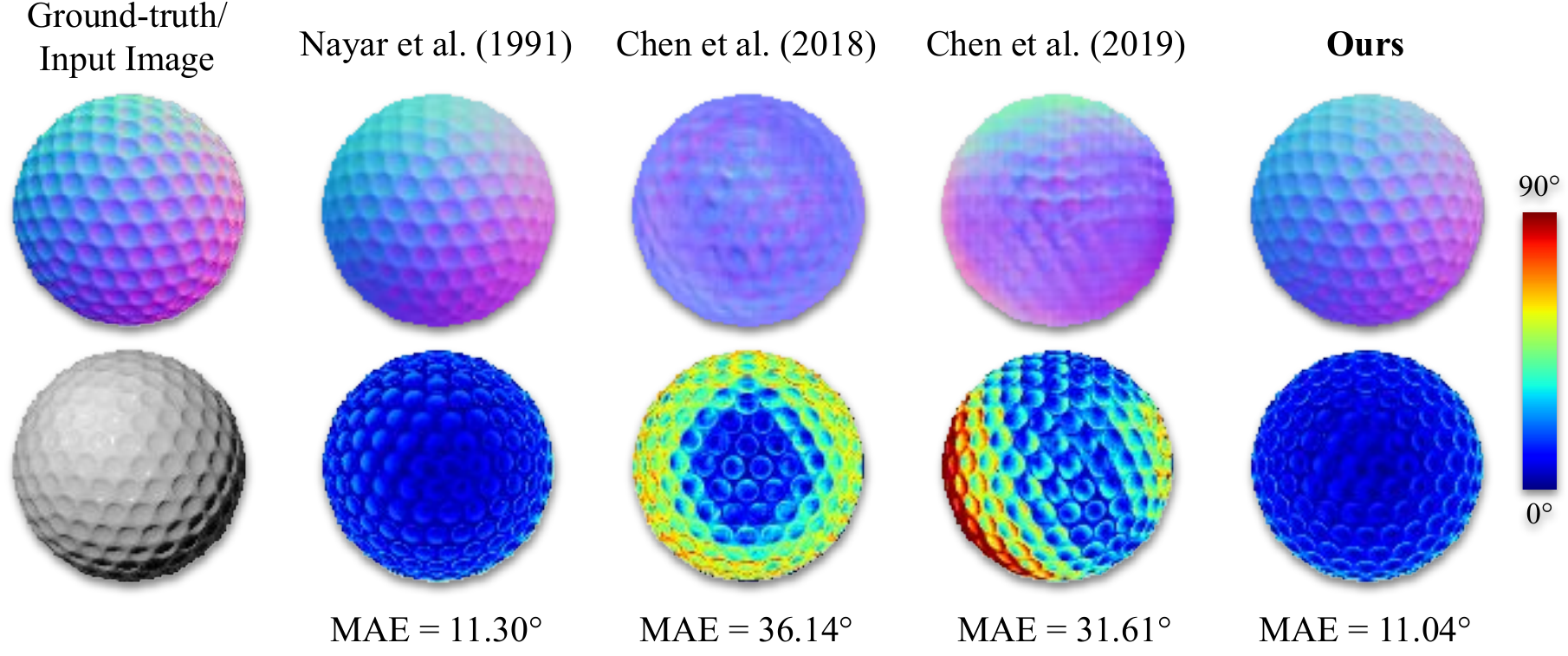}
\caption{ \footnotesize Qualitative comparison on the \textbf{Golf-ball} scene. Although deep learning based methods perform well smooth objects, they cannot handle fine structures and indentations. 
}
\label{fig:golf}
\end{figure}

\begin{figure}
\centering
    \includegraphics[{width=\linewidth}]{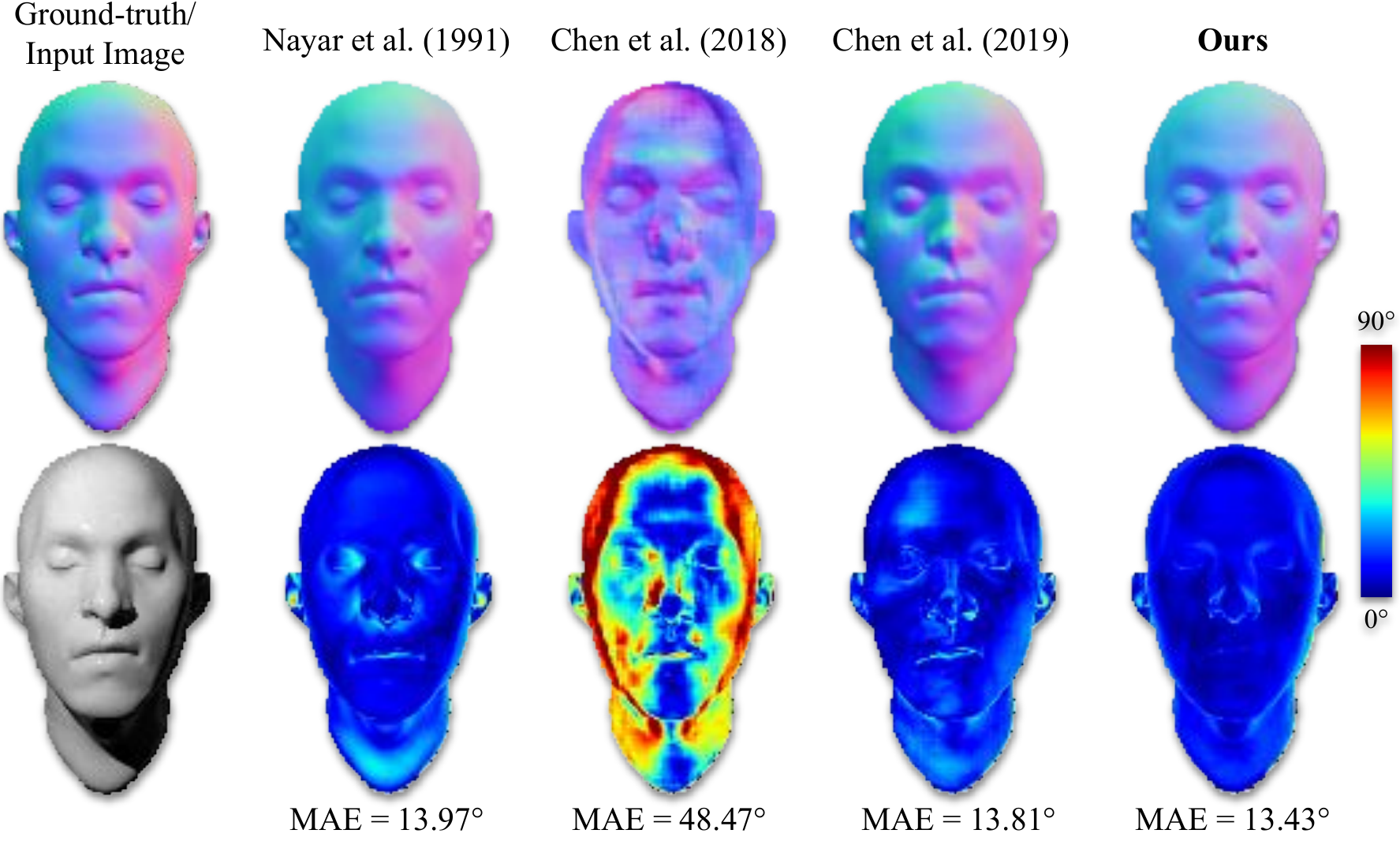}
\caption{ \footnotesize Qualitative comparison on the \textbf{Face} scene. Although Nayar \etal \cite{nayar1991shape} models interreflections, it cannot handle cast shadows. Therefore, it performs poorly on regions surrounding the eyes and the nose where cast shadows are effective. Here, we also observe that Chen \etal \cite{chen2018ps} cannot estimate accurately for higher slant angles. 
}
\label{fig:face}
\end{figure}

\begin{figure}
\centering
\includegraphics[{width=\linewidth}]{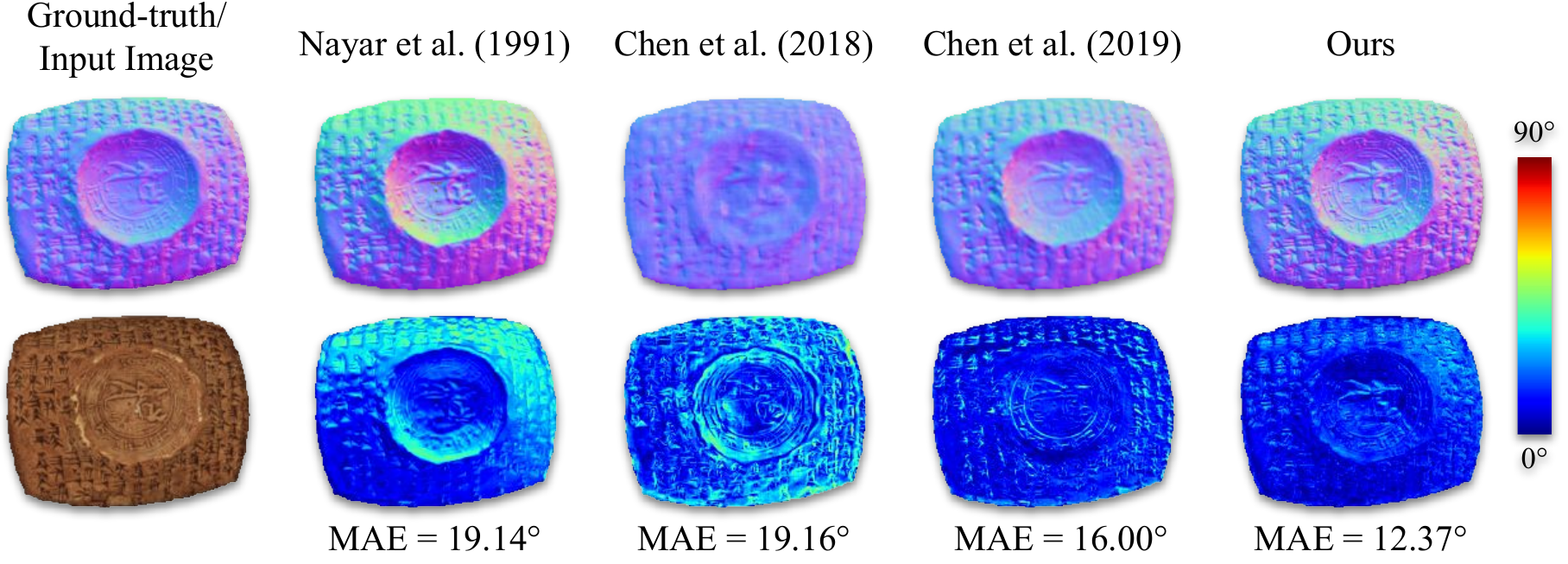}
\caption{\footnotesize Qualitative comparison on the \textbf{Tablet1} scene. This subject has a complicated geometry involving cuneiform and reliefs. Apart from these fine structures, the object can be treated as a composite surface which has a large concavity in the middle part. 
}
\label{fig:tablet1}
\end{figure}

\begin{figure}
\centering
\includegraphics[{width=\linewidth}]{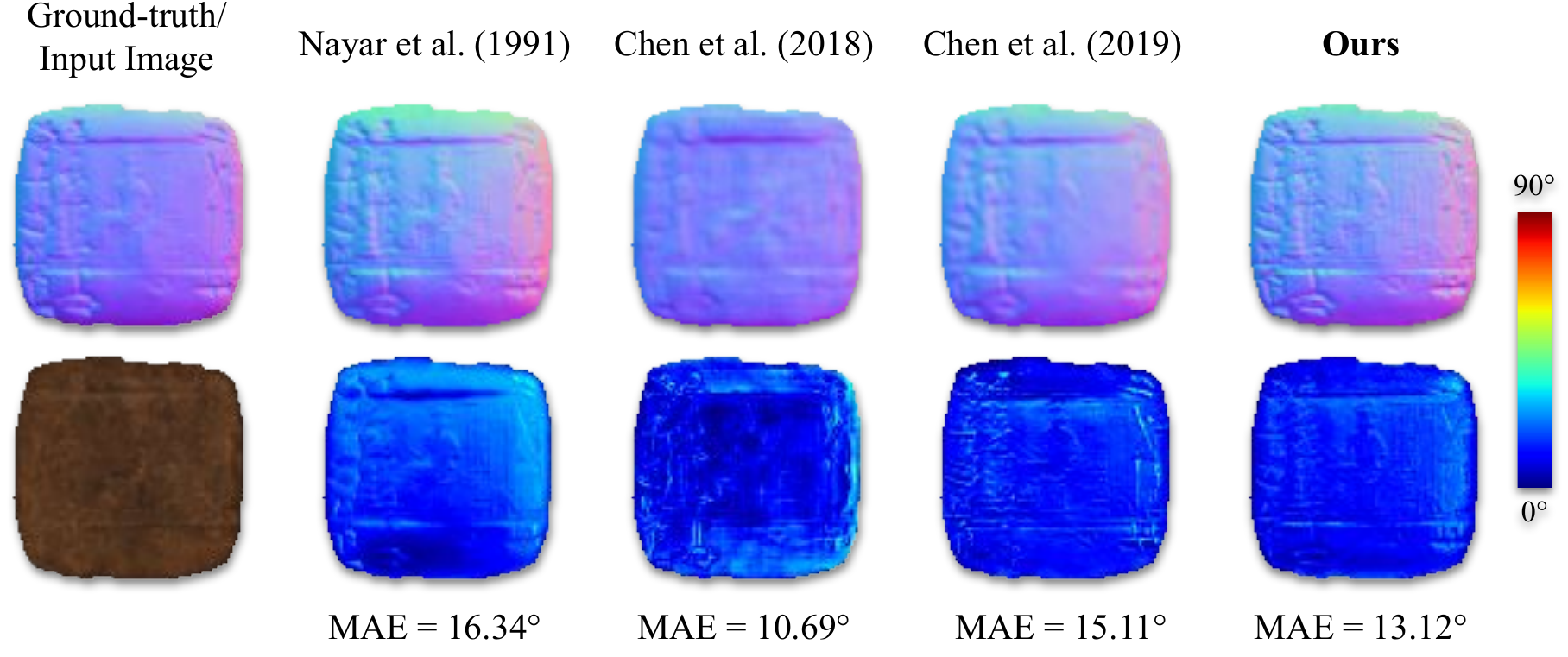}
\caption{ \footnotesize Qualitative comparison on the \textbf{Tablet2} scene. Similar to \textit{Tablet1}, this subject also contains reliefs and cuneiform scripts. Since the overall geometry is approximately flat, all methods perform comparable on this category.
}
\label{fig:tablet2}
\end{figure}

\begin{figure}
\centering
\includegraphics[{width=\linewidth}]{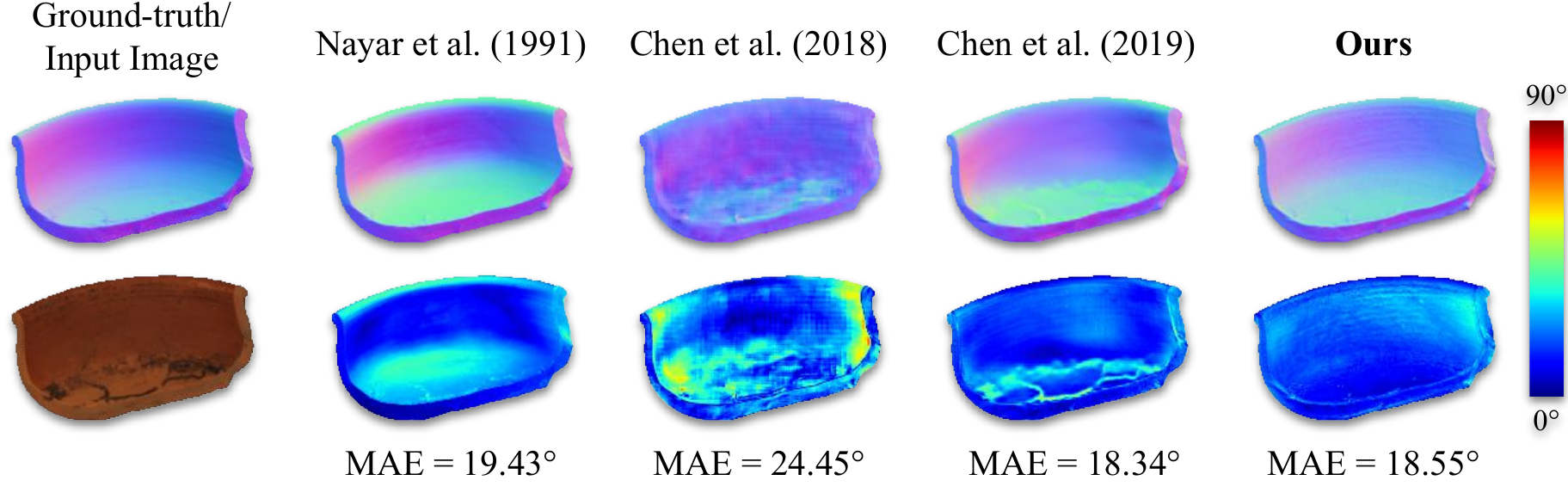}
\caption{ \footnotesize Qualitative comparison on the \textbf{Broken Pot} scene.
}
\label{fig:brokenpot}
\end{figure}

Here, we present qualitative results on all of the categories of our proposed dataset. Figure \ref{fig:vase} to Figure \ref{fig:brokenpot} compares the output normal maps of our method with other baselines. Note that our implementation of Nayar \etal \cite{nayar1991shape} uses Woodham's classical photometric stereo \cite{woodham1980photometric} to calculate the pseudo surface and updates the normals with the interreflection modeling for 15 iterations. Even though the Nayar \etal \cite{nayar1991shape} interreflection algorithm is not theoretically guaranteed to converge for all surfaces, it gives a stable response on our dataset. We initialized Nayar's algorithm using the same predicted light sources of our method for a fair comparison. 

The results show that our method achieves the best results overall, both qualitatively and quantitatively. We observed that other deep learning networks \cite{chen2018ps, chen2019self} may fail to remove the surface ambiguity in challenging subjects. This is because these networks require supervised training with ground-truth normals, and their performance depends on the content of the training dataset. On the other hand, the results show that Nayar \etal \cite{nayar1991shape} performs much better on challenging concave shapes. However, it cannot model specularities and cast shadows. On the other hand, our method can model these non-Lambertian effects with the reflectance mapping, and therefore, it performs better than Nayar \etal. in all the tested categories.

\begin{figure}
\centering
    \includegraphics[{width=\linewidth}]{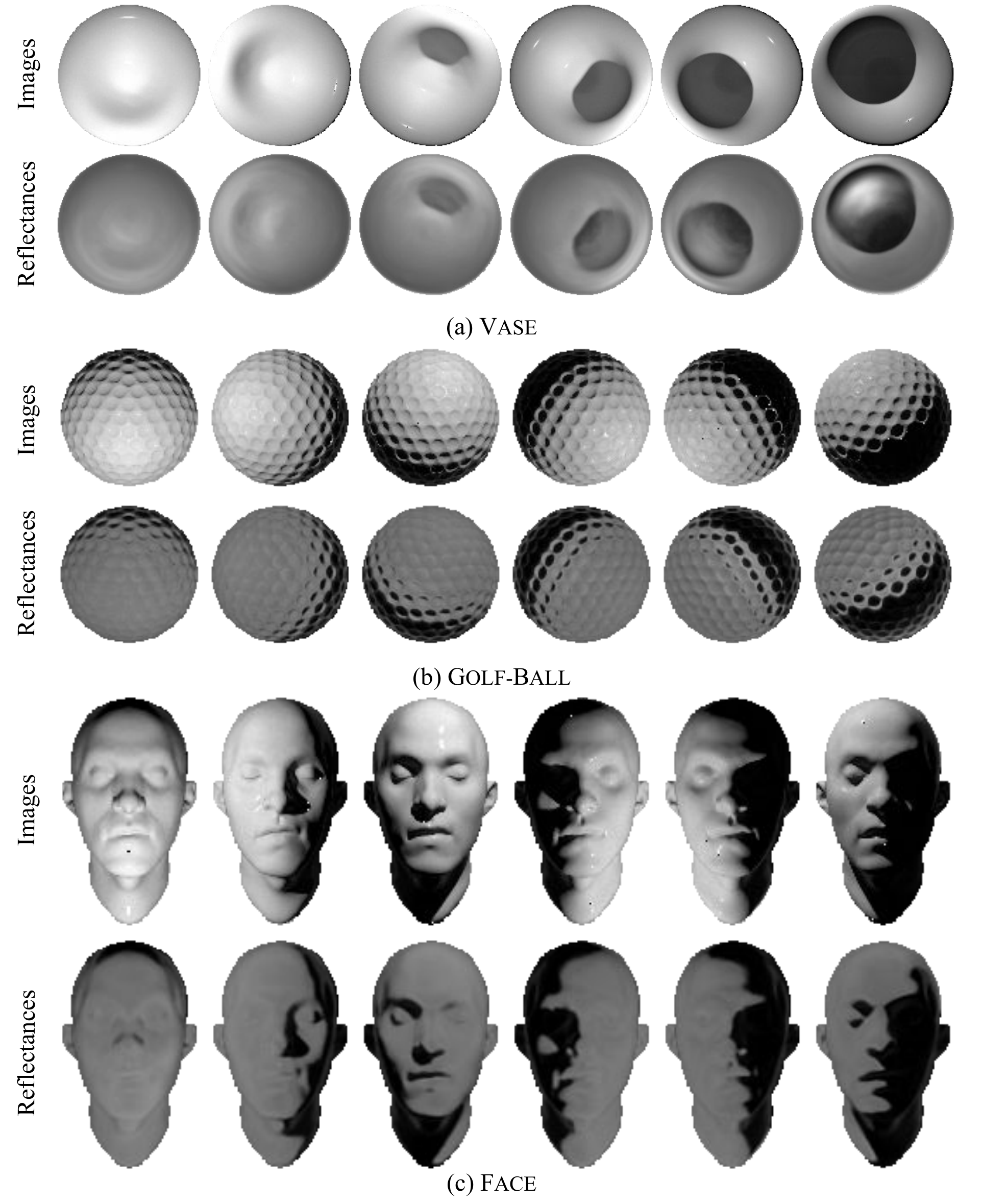}
\caption{ \footnotesize Reflectance maps obtained with our method from \textbf{Vase}, \textbf{Golf-ball} and \textbf{Face} categories. 
}
\label{fig:reflectances_synthetic}
\end{figure}

\begin{figure}
\centering
    \includegraphics[{width=\linewidth}]{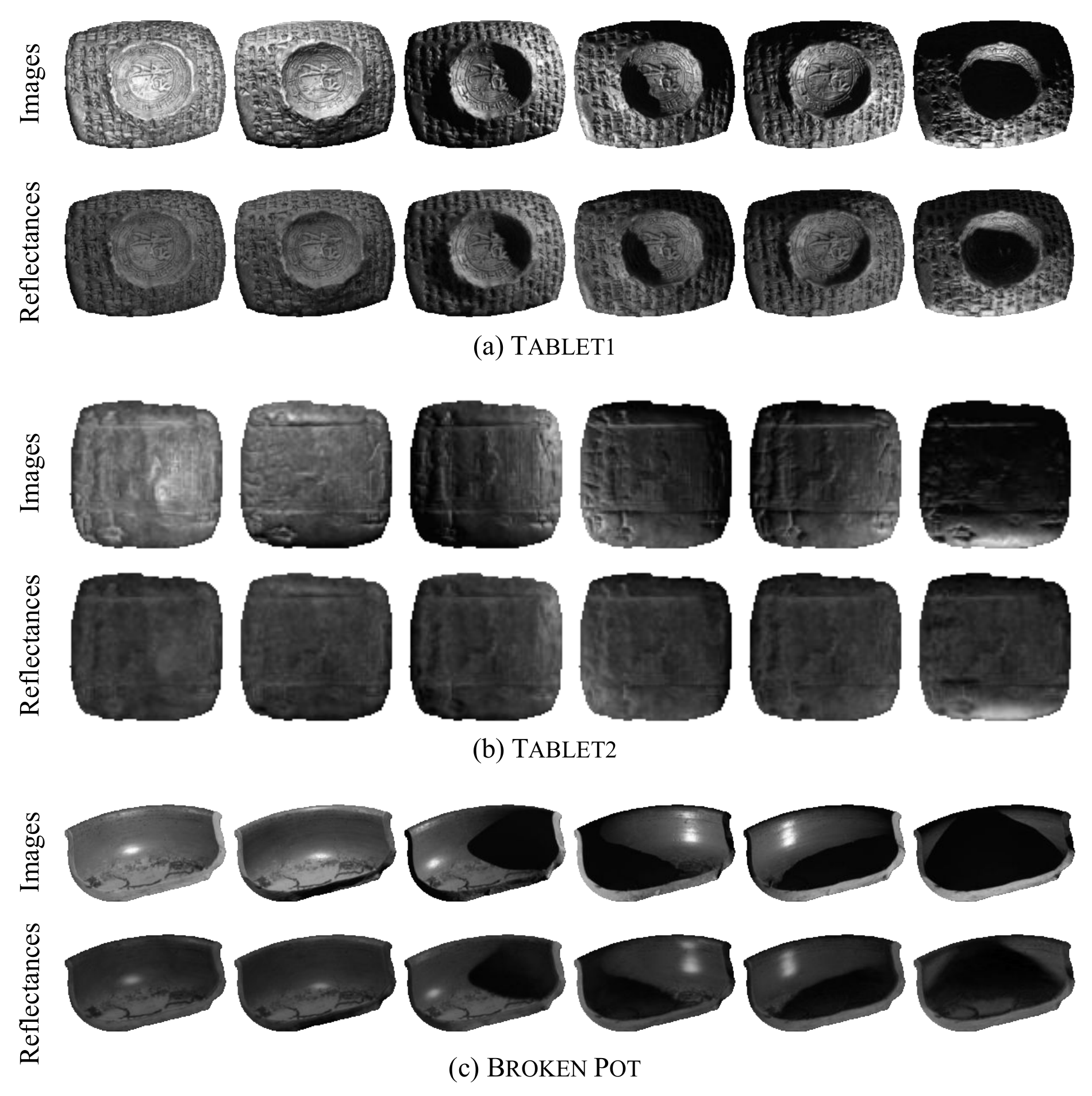}
\caption{ \footnotesize Reflectance maps obtained with our method from \textbf{Tablet1}, \textbf{Tablet2} and \textbf{Broken Pot} categories. 
}
\label{fig:reflectances_real}
\end{figure}

Lastly, we provide the reflectance map obtained using our method on the proposed dataset. Figure \ref{fig:reflectances_synthetic} and Figure \ref{fig:reflectances_real} show the reflectance map obtained using our method on the synthetic and real sequence respectively.

\section{Some General Comments}

\textit{\textbf{Q1}: Influence of complex texture on the light estimation}.
Indeed, surface texture is can be important for light estimation. However, the present benchmark datasets \emph{i.e.}, DiLiGenT is composed of textureless subjects, and therefore, our focus was to perform surface reconstruction on textureless objects.

\textit{\textbf{Q2}: Nayar interreflection model vs. Monte Carlo:} Monte Carlo method can provide more photo-realistic renderings. However, such an approach is again expensive, requires analytic BRDF models, and a sophisticated sampling strategy for computation, which can make the pipeline better, but more involved. So, we favored Nayar's method and used reflectance maps to handle non-Lambertian effects.

\begin{table}
\centering
\resizebox{\columnwidth}{!}
{
\begin{tabular}{|c|c|c|c|c|c|c|c|}
\hline
 \rowcolor[gray]{0.85} Ball& Cat  & Pot1 & Bear & Pot2 & Buddha & Goblet & Reading      \\
 \hline
 985 & 2808 & 3601 & 2585 & 2193 & 2787 & 1636 & 1723   \\
 \hline
 \rowcolor[gray]{0.85} Cow & Harvest & Vase & Golf-ball & Face  & Tablet 1 & Tablet 2 & Broken Pot \\
 \hline
 1651 & 3582 & 1280 & 468 &  435 & 1610 & 437 & 1046  \\
 \hline
\end{tabular}
}
\caption{\footnotesize Number of facets per subject used for our experiments.}
\label{tab:facet_details}
\end{table}

\noindent
\textit{\textbf{Q3}: Number of parameters for the normal estimation network and interreflection kernel computation:}
The inverse rendering network has $\approx$3.7 million parameters (12.3 MB). The interreflection kernel is generally sparse, and efficient software are available to handle large-sized sparse matrices. Table (\ref{tab:facet_details}) provides the number of facets used for our experiments to calculate the interreflection kernel.

\section{Other Possible Future Extension}
Our proposed method enables the application of photometric stereo on a broader range of objects. Yet, we think that there are possible future directions to extend it. Firstly, our method is generally a two-stage framework that utilizes a light estimation network and inverse rendering network in separate phases during inference.  As an extension of our work, we aim to combine those stages in an end-to-end framework where light, surface normals, and reflectance values are estimated simultaneously. Secondly, our method uses a physical rendering equation for image reconstruction that is not sufficient for modeling all physical interactions between the object and the light. We believe that an improved rendering equation with additional physical constraints will allow better normal estimates. In addition to that, our method utilizes a specular-reflectance map inspired by the Phong reflectance model. Using other sophisticated variants of specular-reflectance map such as the Blinn-Phong reflection model \cite{blinn1977models} may further advance our approach.  Finally, we observed that our method is very convenient for practical usage as it doesn't require ground-truth normals for supervised training. However, it could be possible to improve performance by utilizing training data in a similar framework.

\end{document}